\PassOptionsToPackage{table,xcdraw}{xcolor}
\documentclass[10pt,twocolumn,letterpaper]{article}
\usepackage[pagenumbers]{cvpr}      

\definecolor{cvprblue}{rgb}{0.21,0.49,0.74}
\usepackage[pagebackref,breaklinks,colorlinks,citecolor=cvprblue]{hyperref}
\usepackage{graphicx}
\usepackage{amsmath}
\usepackage{booktabs}
\usepackage{bm}
\usepackage{makecell}
\usepackage{multirow}
\usepackage{algorithm}
\usepackage{algorithmic}
\usepackage[flushleft]{threeparttable}
\usepackage{url}
\usepackage{bbding}
\usepackage{utfsym}
\usepackage{float}
\usepackage{subfloat}
\usepackage{subcaption}
\usepackage{amssymb}
\usepackage{arydshln}
\usepackage{colortbl}
\usepackage[T1]{fontenc}

\makeatletter

\newcommand{\Rmnum}[1]{\expandafter\@slowromancap\romannumeral #1@}
\makeatother

\title{Efficient Video Face Enhancement with Enhanced Spatial-Temporal Consistency}

\author{
Yutong Wang$^{1}$\quad Jiajie Teng$^{2}$\quad Jiajiong Cao\quad Yuming Li\quad Chenguang Ma\quad Hongteng Xu$^{3}$ \\ Dixin Luo$^{1}$\thanks{Corresponding author.}\\
$^1$Beijing Institute of Technology \quad 
$^2$Zhejiang University \quad 
$^3$Renmin University of China\\
{\tt\small \{yutongwang1012, dxluo611\}@gmail.com}
}

\begin{document}
\maketitle
\begin{abstract}
As a very common type of video, face videos often appear in movies, talk shows, live broadcasts, and other scenes. 
Real-world online videos are often plagued by degradations such as blurring and quantization noise, due to the high compression ratio caused by high communication costs and limited transmission bandwidth. 
These degradations have a particularly serious impact on face videos because the human visual system is highly sensitive to facial details.
Despite the significant advancement in video face enhancement, current methods still suffer from $i)$ long processing time and $ii)$ inconsistent spatial-temporal visual effects (e.g., flickering). 
This study proposes a novel and efficient blind video face enhancement method to overcome the above two challenges, restoring high-quality videos from their compressed low-quality versions with an effective de-flickering mechanism. 
In particular, the proposed method develops upon a 3D-VQGAN backbone associated with spatial-temporal codebooks recording high-quality portrait features and residual-based temporal information. 
We develop a two-stage learning framework for the model. 
In Stage \Rmnum{1}, we learn the model with a regularizer mitigating the codebook collapse problem.
In Stage \Rmnum{2}, we learn two transformers to lookup code from the codebooks and further update the encoder of low-quality videos.
Experiments conducted on the VFHQ-Test dataset demonstrate that our method surpasses the current state-of-the-art blind face video restoration and de-flickering methods on both efficiency and effectiveness.
Code is available at \url{https://github.com/Dixin-Lab/BFVR-STC}.
\end{abstract}

\section{Introduction}

Real-world videos usually suffer from varying degradations due to shooting methods, transmission compression, etc., which leads to blurring, noise, brightness flickering, color jitter, and other perturbations. 
With the development of video generation technologies such as Sora~\cite{openai2024sora}, AI-generated videos have few problems with content quality. 
Still, the inherent randomness of the generative model often leads to inconsistency between frames thus causing AI-based (pixel) flickering, resulting in its limited practical applications. 
In recent years, deep learning techniques have greatly contributed to the development of the computer vision field, giving rise to a series of works dedicated to achieving video enhancement, including video restoration~\cite{wang2019edvr,liang2024vrt}, video super-resolution~\cite{kappeler2016video,shi2016real}, video de-flickering~\cite{lei2023blind,fulari2024unsupervised}, video colorization~\cite{zhao2023svcnet,yatziv2006fast}, and so on.

The face usually serves as a common focal point or region of interest in media such as images and videos, and a huge amount of research has been devoted to realizing face restoration. 
Blind face restoration (BFR) aims to reconstruct high-quality faces from low-quality faces suffering from multiple degradations~\cite{li2018learning}. 
A great number of excellent works have already been proposed to address blind face image restoration (BFIR) from the perspective of face geometry priors~\cite{chen2021progressive,chen2018fsrnet,hu2020face}, generative priors~\cite{karras2019style,karras2020analyzing,chan2021glean,yang2021gan}, and codebook priors~\cite{wang2022restoreformer,gu2022vqfr,zhou2022towards}. 
However, applying BFIR methods directly to degraded videos frame-by-frame will lead to inconsistencies in background and portrait features between frames due to the lack of temporal constraints. 
Hence, works on blind face video restoration (BFVR) often adopt carefully crafted strategies to enhance the correlation between frames~\cite{xu2024beyond,feng2024kalman,tan2024blind}. 
Still, these works suffer from inefficiency, e.g., long inference times, and an inadequate temporal perceptive field. 
KEEP~\cite{feng2024kalman} needs an additional restoration method RealESRGAN~\cite{wang2021real} and face detection model to process background and faces separately, 
and StableBFVR~\cite{tan2024blind} requires the restoration method BasicVSR++~\cite{chan2022basicvsr++} for preliminary restoration. 
The perceptive field of PGTFormer~\cite{xu2024beyond} considers only two neighboring frames, which makes it difficult to ensure global consistency.

To improve the above BFVR paradigms, in this study, we propose a novel 3D-VQGAN framework to enable efficient video-level compression and quantization. 
To conquer the instability and artifacts caused by video-level compression, we leverage a pre-trained feature network, such as DINOv2, as a more powerful discriminator, followed by multiple multi-scale discriminator heads. 
In terms of quantization, the intrinsic properties of videos motivated us to design spatial-temporal codebooks, where the spatial codebook is used to record high-quality portrait features and the temporal codebook to record residual-based temporal information. 
We also propose a marginal prior regularization to alleviate the codebook collapse problem caused by nearest neighbor retrieval as well as multiple codebooks.

The overall video face enhancement framework comprises two stages. 
The Stage \Rmnum{1} trains the spatial-temporal codebooks and high-quality (HQ) auto-encoders by reconstruction. 
The Stage \Rmnum{2} predicts the code sequences of spatial-temporal compressed tokens in two codebooks by two code lookup transformers respectively. 
The training of Stage \Rmnum{2} fixes the codebooks and decoder learned in Stage \Rmnum{1} and trains the code lookup modules and low-quality (LQ) encoder by HQ-LQ video pairs. 
The proposed enhancement framework can be used to remove degradations such as low resolution, noise, and blurring in BFR tasks, and can also be used to remove brightness flickering in real-world videos and flickering due to intra-frame inconsistencies in AI-generated videos.

In summary, the contribution of our work includes: 
\begin{itemize}
    \item We propose a novel video face enhancement framework that efficiently solves the BFVR and de-flickering tasks. 
    \item We propose a 3D-VQGAN backbone and a powerful discriminator that can support effective video compression. 
    \item We propose spatial-temporal codebooks for video quantization and marginal prior regularization to mitigate the codebook collapse problem. 
    \item We conduct comprehensive experiments on VFHQ-Test on BFVR and de-flickering tasks and compare ours to state-of-the-art methods. Experimental results demonstrate the effectiveness of our method. 
\end{itemize}

\section{Related Works}

\subsection{Blind Face Restoration}
Early blind face image restoration (BFIR) methods usually employ geometry priors-based methods~\cite{chen2021progressive,chen2018fsrnet,hu2020face}. 
However, for highly-degraded images, the geometry information such as face landmarks and face heatmaps will differ drastically from that of the real images. 
Subsequently, researchers attempt to introduce the power of generative models as the generative prior for better visual effects and model robustness. 
The works in~\cite{karras2020analyzing,chan2021glean,yang2021gan} adopt the pre-trained StyleGAN~\cite{karras2019style} as the generator, while the works in~\cite{lin2023diffbir} leverage pre-trained Stable Diffusion as prior information to achieve realistic restoration.  
The works in~\cite{wang2022restoreformer,gu2022vqfr,zhou2022towards} propose to train a VQGAN with a codebook to learn discrete high-quality portrait features, which can be seen as a special case of generative prior. 
On the basis of BFIR methods, blind face video restoration (BFVR) methods focus more on the temporal consistency between frames. 
PGTFormer~\cite{xu2024beyond} proposes to derive face parsing context cues from input video to reduce artifacts and jitters. 
StableBFVR~\cite{tan2024blind} inserts temporal layers in the diffusion process to incorporate temporal information into the generative prior. 
KEEP~\cite{feng2024kalman} tries to leverage previously restored frames to guide the restoration of the current frame by Kalman filtering principles.

\subsection{Video De-flickering}
Video flickering is common in real-world and AI-generated videos, resulting in poor visual viewing. 
The reasoning for flickering in two types of video is different. 
In real-world videos, flickering usually comes from the shooting equipment and environment. 
For example, the unstable brightness flickering in old films is usually due to outdated cameras that can't set a fixed exposure time~\cite{delon2010stabilization}. 
Prior works require additional guidance to remove the flicker, such as flickering frequency~\cite{kanj2017flicker} and counterpart unprocessed video~\cite{bonneel2015blind}.
In contrast, pixels from inconsistent texture and details between frames are responsible for flickering in AI-generated videos. 
FastBlend~\cite{duan2023fastblend,duan2024diffutoon} adopts intra-frame fuzzy interpolation to solve flickering in AI-generated videos, but the generated results suffer from detail loss and blurring. 
The work in~\cite{lei2023blind} utilizes the neural atlas with a neural filtering strategy to learn consistent features and provide temporal guidance. 
Our proposed video face enhancement framework can effectively and efficiently conquer both the brightness flickering in real-world face videos and pixel flickering in AI-generated face videos.

\subsection{Codebook Collapse}
Vector quantization (VQ) quantizes continuous feature vectors into a discrete space by projecting them to the closest items in a codebook, which has been widely used in multiple vision and language tasks~\cite{razavi2019generating,rombach2022high,zheng2022movq}. 
However, optimizing the code items in vector quantization models (e.g., VQ-VAE~\cite{van2017neural} and VQ-GAN~\cite{esser2021taming}) is not toilless. 
The nearest neighbor retrieval method~\cite{zhang2023regularized} and the stop-gradient operation caused by quantization~\cite{huh2023straightening} lead to the codebook collapse problem, which means only a small subset of items receive gradients for optimization while a majority of them are never updated. 
A strenuous effort has been made to alleviate codebook collapse, such as Exponential Moving Averages (EMA)~\cite{van2017neural}, codebook reset~\cite{lancucki2020robust,zeghidour2021soundstream}, probabilistic sampling~\cite{takida2022sq,baevski2019vq}, and anchor updating~\cite{zheng2023online}. 
The work in~\cite{zhang2023regularized} proposed to regularize the tokens used for quantization with a pre-defined uniform distribution. 
However, it uses retrieved indexes to count the frequency of each code item, which is still unfair to inactive code items. 
We propose marginal prior regularization, which counts the frequency of code items by accumulating similarity scores, instead of one-hot encodings, further improving the utilization of codebooks. 

\begin{figure*}[t]
    \begin{center}
    \includegraphics[width=1.0\textwidth]{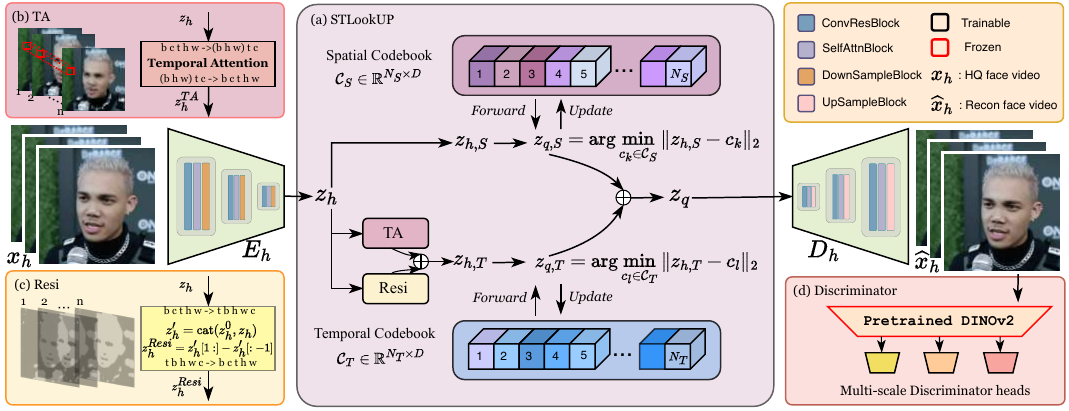}
    \end{center}
    \caption{\textbf{Network architecture of Stage I.} 
    Stage I uses HQ face videos to train \textbf{HQ 3D-VQGAN} ($E_h$ and $D_h$) and \textbf{spatial and temporal codebooks} ($\mathcal{C}_S$ and $\mathcal{C}_T$).
     \textbf{(a)} illustrates the quantization operation STLookUp through two codebooks in our proposed framework. \textbf{(b)} and \textbf{(c)} display the computation process of temporal attention and motion residual, respectively. 
     \textbf{(d)} We leverage a pre-trained feature network DINOv2 and trainable multi-scale discriminator heads to construct a more powerful discriminator for stable training. }
    \label{fig:stage1_shcema}
\end{figure*}

\section{Methodology}

\subsection{Overview}

\begin{figure*}[t]
    \begin{center}
    \includegraphics[width=1.0\textwidth]{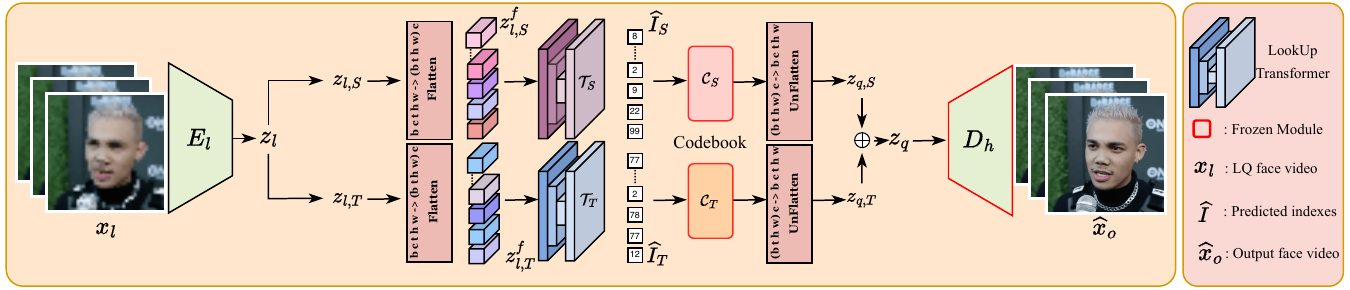}
    \end{center}
    \caption{\textbf{Network architecture of Stage \Rmnum{2}.}
    Stage \Rmnum{2} uses HQ-LQ face video pairs to train LQ encoder $E_l$ and LookUp Transformers ($\mathcal{T}_S$ and $\mathcal{T}_T$). The weights of $D_h$ are pre-trained in Stage \Rmnum{1} and fixed in Stage \Rmnum{2}.
    }
    \label{fig:stage2_shcema}
\end{figure*}

The primary objective of video face enhancement is to reconstruct high-quality (HQ) face videos from heavily-degraded low-quality (LQ) face videos, which may suffer from downsampling, blurring, noise, flickering, and color jitters. 
The limitations of existing video face enhancement methods are the inefficiencies arising from long processing chains and the limited temporal perceptive field that cannot guarantee global consistency. 
We extend the VQGAN paradigm to the video domain to more effectively capture spatial-temporal information for a more direct and comprehensive video face enhancement. 
The proposed framework comprises two stages, as illustrated in Fig~\ref{fig:stage1_shcema} and Fig~\ref{fig:stage2_shcema}. 
In Stage \Rmnum{1}, we utilize HQ videos to train temporal and spatial codebooks under a 3D-VQGAN backbone, through which the discrete HQ facial representations and temporal shifts can be stored in the codebooks in a self-supervised manner. 
In Stage \Rmnum{2}, using HQ-LQ video pairs, we incorporate two transformers to predict temporal and spatial code indices of patches from LQ inputs.

\subsection{Codebook Learning (Stage \Rmnum{1})}

In Stage \Rmnum{1}, we utilize HQ face videos to pre-train a convolution-based 3D-VQGAN model and spatial-temporal codebooks to learn HQ facial representations and capture temporal dynamics. 
We design a more powerful discriminator to stabilize model training and reduce artifacts. 
Besides, we introduce a marginal prior regularization to mitigate the issue of codebook collapse. 

\noindent\textbf{3D-VQGAN.}
The input consists of a sequence of HQ face video frames denoted as $\bm{x}_{h}\in\mathbb{R}^{T\times H\times W\times 3}$. 
These frames are fed into the 3D encoder $E_h$, which produces spatial-temporal compressed latent representations $\bm{z}_h\in\mathbb{R}^{t\times h\times w\times D}$. 
With meticulously designed spatial-temporal codebooks $\bm{\mathcal{C}}_S$ and $\bm{\mathcal{C}}_T$, we derive the quantized latent representations $\bm{z}_q\in\mathbb{R}^{t\times h\times w\times D}$ through nearest neighbor (NN) codebook retrieval. 
Finally, the 3D decoder $D_h$ reconstructs the quantized representations into the output $\widehat{\bm{x}}_{h}\in\mathbb{R}^{T\times H\times W\times 3}$.
\begin{eqnarray}\label{eqn:vae}
\begin{aligned}
    \bm{z}_h&=E_h(\bm{x}_h;\theta_{E_h}), \\
    \bm{z}_q&=\text{STLookUp}(\bm{z}_h;\bm{\mathcal{C}}_S,\bm{\mathcal{C}}_T), \\
    \widehat{\bm{x}}_h&=D_h(\bm{z}_q;\theta_{D_h}).
\end{aligned}
\end{eqnarray}

The encoder and decoder are implemented as purely convolutional structures to support faster processing efficiency while accommodating inputs of various resolutions. 
The autoencoder is constructed by multiple blocks, each containing a residual block, a down/up-sampling block, and a convolution-based self-attention block. 
Spatial-temporal compression is achieved through downsampling layers, where the degree of spatial compression is larger than that of temporal compression so that both spatial-only and spatial-temporal versions of the sampling layer will exist. 
Guided by the design strategy in prior work~\cite{pku_yuan_lab_and_tuzhan_ai_etc_2024_10948109}, we didn't consider the temporal-only sampling layer. 

Due to the instability and artifact-prone nature of training video-level VQGANs, we propose to leverage a partially initialized and more powerful discriminator. 
Specifically, the discriminator consists of a frozen pre-trained feature network $\mathcal{F}$, such as DINOv2~\cite{oquab2023dinov2,sauer2023adversarial,sauer2024fast}, paired with a set of trainable lightweight discriminator heads $\mathcal{D}_{\phi,k}$. 
For a reconstructed video $\widehat{\bm{x}}_h(\theta)$, the adversarial loss of the model incorporates the outputs from multiple discriminator heads, such as 
\begin{eqnarray}
\begin{aligned}
D_{\phi}\left(\widehat{\bm{x}}_h(\theta)\right)&=-\mathbb{E}_{\bm{x}_h}\left(\sum_{k}\mathcal{D}_{\phi,k}\left(\mathcal{F}\left(\widehat{\bm{x}}_h(\theta)\right)\right)\right), \\
\mathcal{L}_{adv}\left(\bm{x}_h, \theta, \phi\right)&=[\log D_{\phi}(\bm{x}_h)+\log (1-D_{\phi}(\widehat{\bm{x}}_h(\theta)))].
\end{aligned}
\end{eqnarray}

\noindent\textbf{Spatial-temporal codebooks.}
Compared to image face enhancement, the difficulty of video face enhancement lies in maintaining the consistency of facial features and background details between frames. 
The discretization nature of a codebook dictates that applying it directly to video tasks often results in poor temporal consistency, leading to the common flickering problem. 
In addition, conventional codebooks are limited to capturing spatial features and fail to account for the motion information embedded in the videos. 
To tackle the above challenges, we propose learnable motion-aware spatial-temporal codebooks, where the spatial codebook $\bm{\mathcal{C}}_S=\{\bm{c}_k\in\mathbb{R}^{D}\}_{k=0}^{N_S}$ records portrait features and the temporal codebook $\bm{\mathcal{C}}_T=\{\bm{c}_l\in\mathbb{R}^{D}\}_{l=0}^{N_T}$ stores motion residuals between frames.

Given the compressed latent representation $\bm{z}_h\in\mathbb{R}^{t\times h\times w\times D}$ output by $E_h$, we first calculate the spatial and temporal latents, respectively. 
The spatial latents $\bm{z}_{h,S}$ are directly derived from $\bm{z}_h$, whereas the temporal latents $\bm{z}_{h,T}$ incorporate inter-frame temporal-attention (TA) information and motion residuals, defined as follows:
\begin{eqnarray}\label{eqn:divide}
\bm{z}_{h,S}=\bm{z}_h;\quad
\bm{z}_{h,T}=\text{TA}(\bm{z}_h)+\text{Residual}(\bm{z}_h)
\end{eqnarray}
where the motion residuals are defined as the difference between latents of two frames separated by a time window, as illustrated in Fig~\ref{fig:stage1_shcema} (b) and (c). 
Subsequently, we can calculate the code (indices) sequences and quantized latents corresponding to spatial and temporal latents respectively based on nearest neighbor retrieval:
\begin{eqnarray}\label{eqn:retrieval}
\begin{aligned}
I_{S}^{(i,j)}&=\mathop{\arg\min}_{k}\|\bm{z}_{h,S}^{(i,j)}-\bm{c}_k\|_{2}\in\{0,\dots, N_S-1\}, \\
\bm{z}_{q,S}^{(i,j)}&=\mathop{\arg\min}_{\bm{c}_k\in\bm{\mathcal{C}}_S}\|\bm{z}_{h,S}^{(i,j)}-\bm{c}_k\|_{2}\in\mathbb{R}^{D}; \\
I_{T}^{(i,j)}&=\mathop{\arg\min}_{l}\|\bm{z}_{h,T}^{(i,j)}-\bm{c}_l\|_{2}\in\{0,\dots, N_T-1\}, \\
\bm{z}_{q,T}^{(i,j)}&=\mathop{\arg\min}_{\bm{c}_l\in\bm{\mathcal{C}}_T}\|\bm{z}_{h,T}^{(i,j)}-\bm{c}_l\|_{2}\in\mathbb{R}^{D}. 
\end{aligned}
\end{eqnarray}

The quantized latents can be obtained by considering two types of latents. The Eqn.~\ref{eqn:retrieval} and Eqn.~\ref{eqn:summation} correspond to the \verb|STLookUp| operation in Eqn.~\ref{eqn:vae}.
\begin{eqnarray}\label{eqn:summation}
    \bm{z}_{q}=\bm{z}_{q,S}\oplus\bm{z}_{q,T}
\end{eqnarray}
where the fusion operator $\oplus$ is set as element-wise addition by default.

\noindent\textbf{Marginal prior regularization.}
Given the flattened spatial latents $\bm{z}^{f}_{h,S}\in \mathbb{R}^{(t\cdot h\cdot w)\times D}$ and spatial codebook $\bm{\mathcal{C}}_{S}\in \mathbb{R}^{N_S\times D}$, we can calculate the Euclidean distance matrix $\bm{D}_{E}\in \mathbb{R}^{(t\cdot h\cdot w)\times N_{S}}$ between them. 
From this, we derive the similarity scores $\bm{s}=\sigma_{r}(1/\bm{D}_{E})=[s_{i,j}]\in\mathbb{R}^{(t\cdot h\cdot w)\times N_{S}}$, where $\sigma_{r}(\cdot)$ means applying normalization operation to each row of the matrix. 
The posterior distribution is approximated by the marginal distribution of the similarity score matrix $\bm{s}$, e.g., $P_{post}=\sum_{i}^{thw} s_{i,j}=[p_1,p_2,\cdots,p_{N_S}]\in\mathbb{R}^{N_S}$. 
The uniform prior distribution is defined as $P_{prior}=[1/N_S,1/N_S,\cdots,1/N_S]\in\mathbb{R}^{N_S}$. 
We leverage KL-divergence to implement the marginal prior regularization:
\begin{eqnarray}\label{eqn:prior_kl}
\begin{aligned}
    &\mathcal{L}_{KL}^{S}=\text{KL}\left(P_{post},P_{prior}\right)=-\sum_{n}^{N_S}p_n \log\left(\frac{1/N_{S}}{p_n}\right).
\end{aligned}
\end{eqnarray}
Similarly, we can calculate $\mathcal{L}_{KL}^{T}$ between flattened temporal latents $\bm{z}^{f}_{h,T}$ and temporal codebook $\bm{\mathcal{C}}_{T}$.

\noindent\textbf{Training objectives.}
In stage \Rmnum{1}, we adopt the loss functions commonly used in VQGAN training, including reconstruction losses $\mathcal{L}_1$, perceptual loss~\cite{johnson2016perceptual,zhang2018unreasonable} $\mathcal{L}_{per}$, adversarial loss~\cite{esser2021taming} $\mathcal{L}_{adv}$ and code-level loss~\cite{van2017neural,esser2021taming} $\mathcal{L}_{f}$: 
\begin{eqnarray}
\begin{aligned}
&\mathcal{L}_1=\|\bm{x}_h-\widehat{\bm{x}}_h\|_{1}, \quad \mathcal{L}_{per}=\|\Phi(\bm{x}_h)-\Phi(\widehat{\bm{x}}_h)\|_{2}^{2}, \\
&\mathcal{L}_{f}=\|\text{sg}(\bm{z}_h)-\bm{z}_q\|_2^2+\beta \|\bm{z}_h-\text{sg}(\bm{z}_q)\|_2^2, \\
&\mathcal{L}_{\Rmnum{1}}=\mathcal{L}_1+\mathcal{L}_{per}+\mathcal{L}_{f}+(\mathcal{L}_{KL}^{S}+\mathcal{L}_{KL}^{T})+\lambda_{adv}\cdot\mathcal{L}_{adv}
\end{aligned}
\end{eqnarray}
where $\Phi$ denotes the feature extractor of VGG19~\cite{simonyan2015very}, $\text{sg}(\cdot)$ is the stop-gradient operator. 
We hope to jointly optimize the encoder $E_h$ and codebooks $\bm{\mathcal{C}}_S$ and $\bm{\mathcal{C}}_T$ via the code-level loss $\mathcal{L}_{f}$, and $\beta$ controls the relative update rates of the encoder and codebooks. 
We also adopt straight-through gradient estimator~\cite{van2017neural,esser2021taming} to solve the non-differentiable problem of quantization operation.

\subsection{Lookup Transformer Learning (Stage \Rmnum{2})}

The heavy degradations in real-world videos not only destroy the structure and details of the original face but also disrupt the continuity of temporal motions. 
In order to support LQ video input $\bm{x}_l$, we introduce two additional Transformer-based code lookup modules designed to predict the code sequence indices of the compressed spatial latents $\bm{z}_{l,S}$ and temporal latents $\bm{z}_{l,T}$ within the codebooks, respectively. 
In stage \Rmnum{2}, we fix the HQ decoder $D_h$ and spatial-temporal codebooks trained in stage \Rmnum{1}, and train the LQ encoder $E_l$ and code lookup modules via HQ-LQ video pairs. 

Given the compressed latents $\bm{z}_l\in\mathbb{R}^{t\times h\times w\times D}$ output by $E_l$, the spatial latents $\bm{z}_{l,S}$ and temporal latents $\bm{z}_{l,T}$ can be calculated analogously to Eqn. (~\ref{eqn:divide}). 
The spatial latents $\bm{z}_{l,S}\in\mathbb{R}^{t\times h\times w\times D}$ are initially flattened into sequence-shaped latents $\bm{z}_{l,S}^{f}\in\mathbb{R}^{(t\cdot h\cdot w)\times D}$ and then processed by a cascaded Transformer module $\mathcal{T}_S$. 
Learnable position embeddings are inserted into the lookup module to enhance the model's awareness of spatial relative positions. 
The lookup module outputs classification probabilities for each sequence item, which are used to generate code sequences $\widehat{I}_{S}\in\{0,\dots, N_S-1\}^{t\cdot h\cdot w}$. 
Similarly, code sequences $\widehat{I}_{T}\in\{0,\dots, N_T-1\}^{t\cdot h\cdot w}$ for temporal latents $\bm{z}_{l,T}$ can be obtained using the code lookup module $\mathcal{T}_{T}$. 
The proposed 3D-VQGAN achieves a substantial compression ratio (\textit{i.e.}, $\frac{H}{h}=\frac{W}{w}=8$ and $\frac{T}{t}=2$), significantly alleviating the computational burden on the Transformers.

\noindent\textbf{Training objectives.}
In stage \Rmnum{2}, we aim to train the LQ encoder $E_l$ and code lookup modules $\mathcal{T}_S$ and $\mathcal{T}_T$ using HQ-LQ video pairs. 
Following the work~\cite{zhou2022towards}, we consider only cross-entropy prediction loss and code-level loss $\mathcal{L}'_{f}$: 
\begin{eqnarray}
\small
\begin{aligned}
&\mathcal{L}_{CE}^{S}=\sum_{k}^{thw}-I_{S}^{k}\log \left(\widehat{I}_{S}^{k}\right),\quad
\mathcal{L}_{CE}^{T}=\sum_{l}^{thw}-I_{T}^{l}\log \left(\widehat{I}_{T}^{l}\right), \\
&\mathcal{L}'_{f}=\|\bm{z}_l-\text{sg}(\bm{z}_q)\|_2^2, ~\
\mathcal{L}_{\Rmnum{2}}=\mathcal{L}'_{f}+\lambda_{CE}\cdot (\mathcal{L}_{CE}^{S}+\mathcal{L}_{CE}^{T})
\end{aligned}
\end{eqnarray}
where $\widehat{I}_S$ and $\widehat{I}_T$ are the code sequences predicted by code lookup modules, while $I_S$ and $I_T$ are the ground-truth code sequences generated by HQ encoder $E_h$ and frozen spatial-temporal codebooks. 
We use code-level loss $\mathcal{L}_{f}^{'}$ to minimize the distance between compressed latents $\bm{z}_l$ and quantized latents $\bm{z}_q$, 
which can simplify and accelerate the training of Stage \Rmnum{2}.

\section{Experiments}

\subsection{Implementation Details}

\noindent\textbf{Datasets.} 
We train the model on the \textbf{VFHQ}~\cite{xie2022vfhq} and our own collected live broadcast datasets. 
The VFHQ dataset contains 16,000 video sequences in total. 
We filtered VFHQ based on face proportion, face orientation, and the presence of text, and finally used approximately 3,200 videos, the data processing details can be found in the Data Manipulation subsection. 
The collected live broadcast dataset contains about 9,000 videos of people talking. 
We evaluate our method and other baselines on \textbf{VFHQ-Test} dataset. 
VFHQ-Test is composed of 50 high-quality video sequences, we choose the first 24 frames (1s) of each sequence as our test set. 

For the blind face video restoration task, our data degradation method is consistent with other competitors~\cite{xu2024beyond,feng2024kalman,tan2024blind}. 
Specifically, the degradation model is as follows~\cite{gu2022vqfr}: 
\begin{eqnarray}
y=\left\{\left[\left(x \circledast k_\sigma\right)_{\downarrow_r}+n_\delta\right]_{\text{FFMPEG}_{crf}}\right\}_{\uparrow_r}
\end{eqnarray}
where $y$ is a degraded frame, $x$ is a HQ frame. $x$ is first convolved with a gaussian kernel $k_\sigma$ to add blurring, then downsample to scale $r$. 
Then, gaussian noise $n_\delta$ and video coding degradation FFMPEG with a constant rate factor are applied into the frames. 
Finally, the output is resized to 512$^2$. 
All frames in one video have a consistent $\sigma$, $r$, and $\delta$.  

For the de-flickering task, we consider brightness flickering and pixel flickering. 
Each frame in the input video has a 30\% probability of being degraded. 
The brightness degradation is implemented through \verb|cv2.convertScaleAbs| to simulate the brightness flickering problem in old movies and damaged videos. 
The pixel flickering is implemented by re-rendering the frame through the Stable Diffusion model to simulate the inconsistency of details and textures between frames.  

\begin{table*}[t]
\centering
\tabcolsep=2.5pt
\begin{tabular}{llccccccccc}
\toprule
\multirow{2}{*}{\textbf{Task}} & \multirow{2}{*}{\textbf{Method}} & \multicolumn{3}{c}{\textbf{Quality and Fidelity}} & \multicolumn{3}{c}{\textbf{Pose Consistency}} & \multicolumn{2}{c}{\textbf{Temporal Consistency}} & \textbf{Efficiency} \\
\cmidrule(lr){3-5} \cmidrule(lr){6-8} \cmidrule(lr){9-10} \cmidrule(lr){11-11}
                          &                         & PSNR$\uparrow$ & SSIM$\uparrow$ & LPIPS$\downarrow$ & AKD$\downarrow$ & Face-Cons$\uparrow$ & IDS$\uparrow$ & FVD$\downarrow$ & Flow-Score$\downarrow$  & Runtime(s)$\downarrow$    \\ \midrule
\multirow{3}{*}{BFIR}     &VQFR~\cite{gu2022vqfr}          &25.94      &0.7852      &0.2467       &5.978      &0.9947       &0.6659   &388.2      &1.451        &15.60                                                          \\ 
                          &GFPGAN~\cite{wang2021towards}        &27.15      &0.8207      &0.2279       &4.134      &0.9950     &0.9206  
    &246.9        &1.316      &14.44                                                          \\
                          &CodeFormer~\cite{zhou2022towards}    &26.77      &0.8102      &0.2373       &4.543      &0.9947           &0.8596      &261.8              &2.672           &28.18                                                          \\ \hline
\multirow{2}{*}{VSR}      &BasicVSR++~\cite{chan2022basicvsr++}    &27.22      &0.8218      &0.2742       &5.129      &\textbf{0.9965}          &0.9234   &392.7                &1.286           &72.21                                                          \\
                          &Real-BasicVSR~\cite{chan2022investigating} &27.45      &0.7929      &0.2968       &4.780           &0.9936     &0.8785     &305.7            &1.404           &12.20                                                          \\ \hline
\multirow{3}{*}{BFVR}     &PGTFormer~\cite{xu2024beyond}     &\textbf{28.68}      &\underline{0.8426}      &\textbf{0.1752}       &\textbf{3.519}      &0.9942          &\underline{0.9296}     &\underline{107.6}            &\underline{1.154}           &\underline{7.085}                                                          \\
                          &KEEP~\cite{feng2024kalman}          &27.04      &0.8223      &0.2370       &3.979      &0.9953         &0.8783     &264.9          &1.302           &19.01                                                          \\ \hdashline
                          &Ours  &\underline{27.47} &\textbf{0.8641} &\underline{0.1829} &\underline{3.858}  &\underline{0.9954}     &\textbf{0.9312}   &\textbf{105.1}  &\textbf{1.150}   &\textbf{2.995} \\ \bottomrule
\end{tabular}
\caption{\textbf{Quantitative comparison on the VFHQ-test dataset for blind face video restoration.} 
The inference efficiency of our method is much higher than that of other baselines. 
We bold the \textbf{best} and underline the \underline{second-best} performances.
} 
\label{tab:cmp1}
\end{table*}

\begin{figure*}[t]
    \centering
    \includegraphics[width=0.9\textwidth]{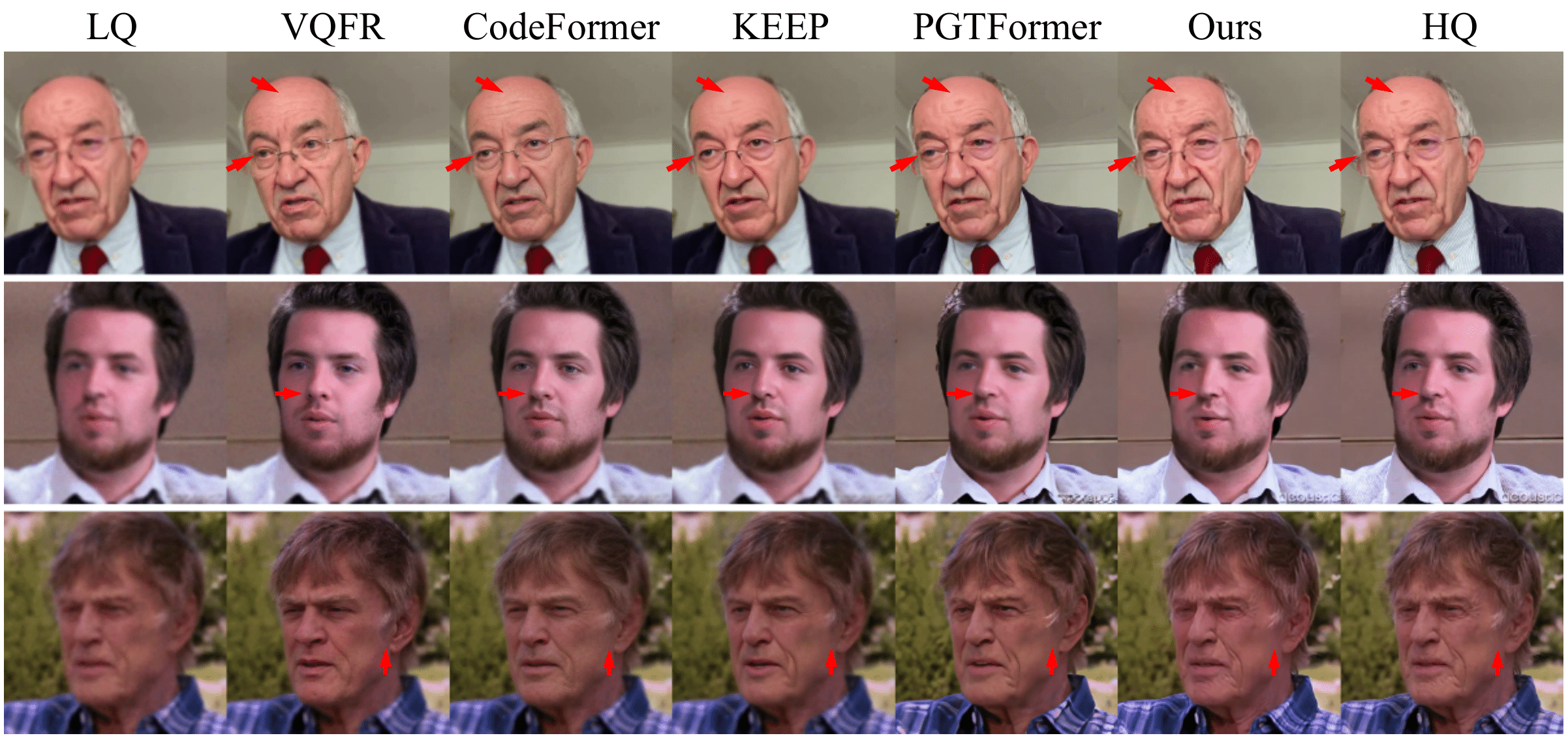}
    \caption{\textbf{Qualitative comparison on the VFHQ-Test for BFVR task.} 
    Our method has better fidelity and fewer hallucination cases compared to other methods, such as wrinkles (1st row), eye orientation (1st row), nose shape (2nd row), and hairstyle (3rd row).
    }
    \label{fig:bfvr_visualizaiton1}
\end{figure*}

\noindent\textbf{Data Manipulation.}
We preprocess the training videos in three aspects, including \textbf{face proportion ($\mathbb{A}$)}, \textbf{face orientation ($\mathbb{B}$)}, and \textbf{the presence of text ($\mathbb{C}$)}. 
Previous methods typically employ a face detection module to separate faces for separate processing, resulting in inefficient inference. 
We improve the quality of the training data by cropping videos based on face proportions, thus simplifying the data processing chains and eliminating the need for face background separation. 
Specifically, we select the region with the largest face proportion in the video as a reference to crop that video to maximize the face region. 
Secondly, the work in~\cite{zhou2022towards} proposes that frontal faces are more conducive to the model learning high-quality portrait features. 
We filtered out side-face-dominated videos by assessing the ratio of the horizontal distances from the eyes to the nose to determine face orientation. 
Thirdly, we found that some text obscures the face in the training data, so we removed videos containing text using PaddleOCR.

\noindent\textbf{Evaluation Metrics.}
For the evaluation of the BFVR task, we adopt \textbf{PSNR}, \textbf{SSIM}, and \textbf{LPIPS}~\cite{zhang2018unreasonable} to evaluate the quality and fidelity of restoration videos. 
Following video evaluation benchmark EvalCrafter~\cite{liu2024evalcrafter}, we utilize two metrics to gauge temporal consistency, including \textbf{FVD} and Average Flow (\textbf{Flow-Score}). 
Flow-Score assesses the average dense flows extracted by RAFT~\cite{teed2020raft} from the generated videos. 
Following the work in~\cite{feng2024kalman}, we apply Face Consistency (\textbf{Face-Cons}), Identity Preservation Scores (\textbf{IDS}), and Average Keypoint Distance (\textbf{AKD}) to evaluate identity preservation and pose consistency. 
AKD is calculated by the average distance between detected face landmarks in the generated and ground-truth video frames. 
Face-Cons calculates the average cosine similarity between the embedding of the first frame $emb(x_1)$ and subsequent frame embeddings $\{emb(x_t)\}_{t=2}^{T}$, focusing on the consistency of human identity in generated videos. 
Additionally, we measure the efficiency of different methods by considering the time required to generate a 24-frame (1-second) video. 
For the evaluation of de-flickering tasks, we adopt FVD and Flow-Score to assess temporal consistency. 

\noindent\textbf{Model Configuration.}
In stage \Rmnum{1}, the training videos are resized to 256$^2$ and only the first 24 frames are used for training. 
The latent dimension $D$ is set to 256. 
Both the size of spatial and temporal codebooks are 1024. 
The training iteration is 250,000. 
The used pre-trained feature network $\mathcal{F}$ is \verb|dinov2_vits14|. 
In stage \Rmnum{2}, the training videos are resized to 512$^2$. 
The training iteration is 50,000. 
The degradation degree $\sigma\in[2,5]$, $r\in[2,4]$, $\delta\in[0,5]$. 
The training is conducted on 4 NVIDIA A100 GPUs, with batch size 4. 
Both the two stages use Adam optimizer~\cite{kingma2014adam}. 
We apply \verb|stable-diffusion-x4-upscaler| to generate pixel (AI-based) flickering.

\subsection{Quantitative and Qualitative Comparisons}

\noindent \textbf{Blind face video restoration.}
Our comparison for BFVR task primarily focuses on existing BFIR, VSR, and BFVR methods, including VQFR~\cite{gu2022vqfr}, GFPGAN~\cite{wang2021towards}, CodeFormer~\cite{zhou2022towards}, BasicVSR++~\cite{chan2022basicvsr++}, Real-BasicVSR~\cite{chan2022investigating}, PGTFormer~\cite{xu2024beyond}, and KEEP~\cite{feng2024kalman}. 
We present the quantitative comparison on the VFHQ-Test in Table~\ref{tab:cmp1}. 
Our method outperforms both BFIR and VSR methods across almost all metrics. 
Compared to image-based methods, our generated results exhibit higher quality, fidelity, and improved temporal consistency. 
Among the BFVR methods, our results are comparable to those of state-of-the-art methods, with enhanced inference efficiency. 
This demonstrates that the proposed 3D-VQGAN and spatial-temporal codebooks enable effective video compression and quantization, offering promising restoration quality coupled with reduced inference costs. 
Figure~\ref{fig:bfvr_visualizaiton1} provides some representative restoration results from different methods. 
Our method successfully captures portrait features in low-quality inputs and achieves higher fidelity. 
In contrast, due to a limited temporal perceptive field, other methods inaccurately alter key features, such as wrinkles, eye orientation (1st row), nose shape (2nd row), and hairstyle (3rd row).

\begin{table}[t]
\centering
\tabcolsep=3pt
\begin{tabular}{lcccc}
\toprule
             \textbf{Method} & GT & FVD$\downarrow$  & Flow-Score$\downarrow$ & Runtime(s)$\downarrow$ \\ \hline
Input &- &494.3 &1.572    &- \\ \hline
FastBlend~\cite{duan2023fastblend}   &$\usym{2713}$ &34.58  &1.574 &\textbf{18.44}            \\
DVP~\cite{lei2020blind} &$\usym{2713}$  &\textbf{14.53}  &\textbf{1.492} &410.2   \\ \hline
NeuralAtlas~\cite{lei2023blind} &$\usym{2717}$ &111.0   &1.468    &851.3  \\ 
Ours  &$\usym{2717}$ &\textbf{100.7}     &\textbf{1.100}    &\textbf{2.934}    \\ \bottomrule
\end{tabular}
\caption{\textbf{Quantitative comparison for brightness de-flickering.}}
\label{tab:cmp2}
\end{table}

\begin{table}[t]
\centering
\tabcolsep=3pt
\begin{tabular}{lcccc}
\toprule
             \textbf{Method} &GT & FVD$\downarrow$ & Flow-Score$\downarrow$ & Runtime(s)$\downarrow$ \\ \hline
Input &- &65.32      &1.702 &-           \\ \hline
FastBlend~\cite{duan2023fastblend} &$\usym{2713}$ &45.80   &1.566    &\textbf{18.60}            \\
DVP~\cite{lei2020blind} &$\usym{2713}$  &\textbf{15.09}    &\textbf{1.505 }   &413.9            \\ \hline
NeuralAtlas~\cite{lei2023blind} &$\usym{2717}$ &110.5   &1.471    &854.1            \\ 
Ours  &$\usym{2717}$  &\textbf{86.88}   &\textbf{1.063}    &\textbf{2.967}   \\ \bottomrule
\end{tabular}
\caption{\textbf{Quantitative comparison pixel de-flickering.}}
\label{tab:cmp3}
\end{table}

\noindent \textbf{De-flickering.}
We report the quantitative results for brightness de-flickering and pixel de-flickering in Table~\ref{tab:cmp2} and Table~\ref{tab:cmp3}, respectively. 
For both tasks, we consider three baseline methods, including FastBlend~\cite{duan2023fastblend}, DVP~\cite{lei2020blind}, and NeuralAtlas~\cite{lei2023blind}. 
It should be noted that FastBlend and DVP achieve long-term temporal consistency by integrating or training on a temporally consistent unprocessed video. 
However, such ground-truth videos are not available in real-world application scenarios. 
Unlike these methods, NeuralAtlas and our approach require only a single flickering video without additional guidance. 
Furthermore, the training-at-test-time strategy used by DVP and NeuralAtlas significantly reduces their inference efficiency. 
Our method demonstrates strong temporal consistency while maintaining high inference efficiency, as indicated by the quantitative results. 

\begin{figure*}[htbp]
    \centering
    \begin{subfigure}[b]{0.48\textwidth}
        \centering
        \includegraphics[width=\textwidth]{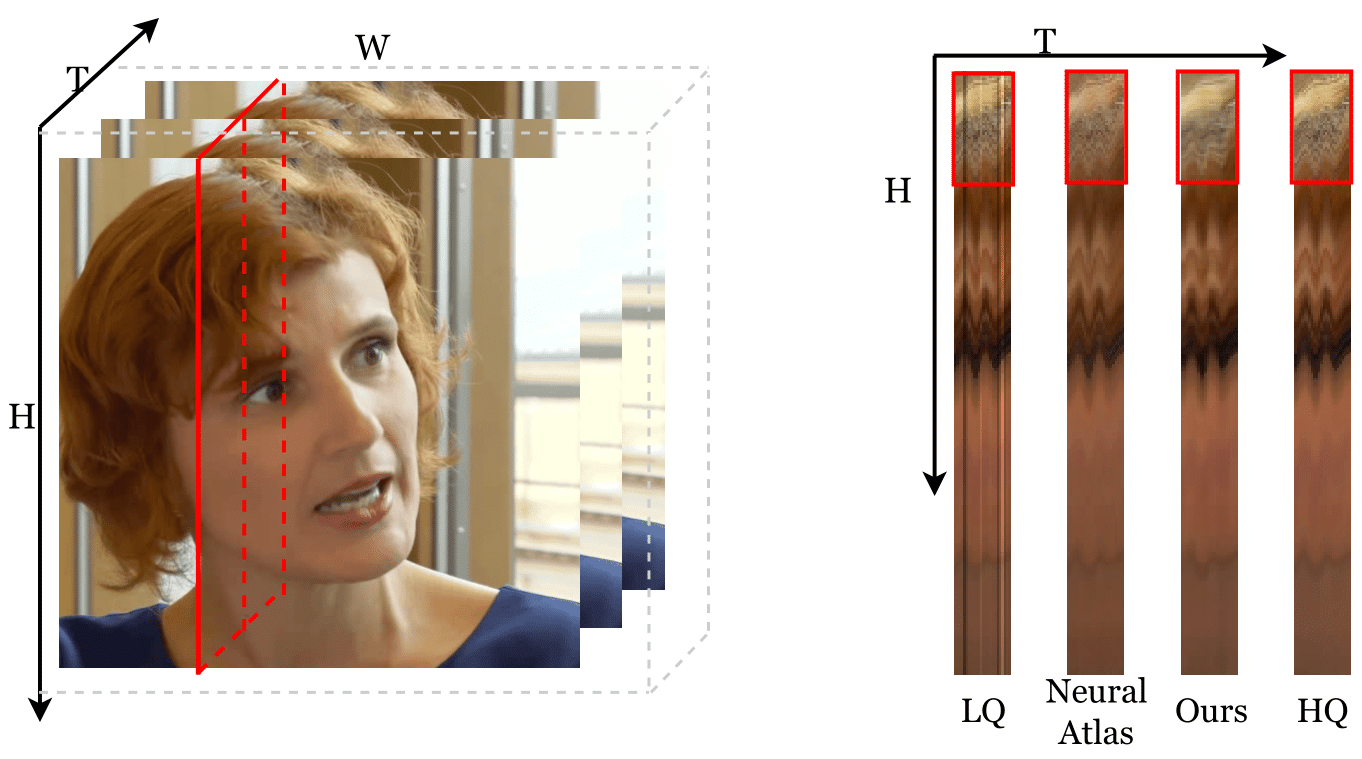}
        \caption{\textbf{Brightness de-flickering.}}
    \end{subfigure}
    \hfill
    \begin{subfigure}[b]{0.48\textwidth}
        \centering
        \includegraphics[width=\textwidth]{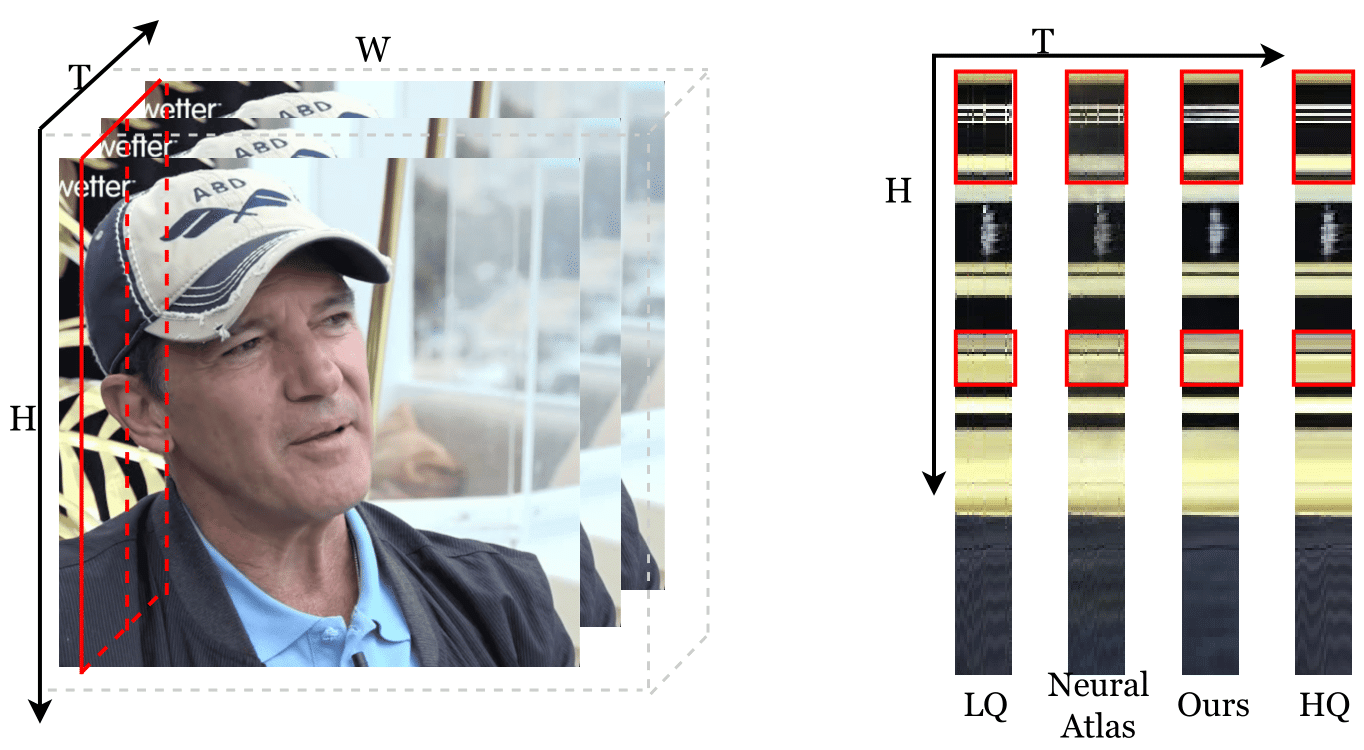}
        \caption{\textbf{Pixel de-flickering.}}
    \end{subfigure}
    \caption{Comparison of temporal profile for brightness and pixel de-flickering. We select a column to observe the changes across time. }
    \label{fig:flicker_visualizaiton1}
\end{figure*}

Figure~\ref{fig:flicker_visualizaiton1} presents the temporal profiles of different methods for brightness de-flickering and pixel de-flickering, respectively. 
We visualize temporal variation by stacking pixels of a particular column over time. 
In Figure~\ref{fig:flicker_visualizaiton1}(a), the LQ input video contains some frames with brightness perturbations, and the generated video of NeuralAtlas is affected by the input resulting in subtle color shifts. 
In contrast, our method, utilizing the 3D-VQGAN backbone and spatial-temporal codebooks, produces de-flickered videos with consistent colors. 
In Figure~\ref{fig:flicker_visualizaiton1}(b), our method better maintains the consistency of the static text and texture in AI-generated videos, effectively addressing common flickering problems. 

\subsection{Ablation Study}

\begin{figure}[t]
    \centering
    \includegraphics[width=0.48\textwidth]{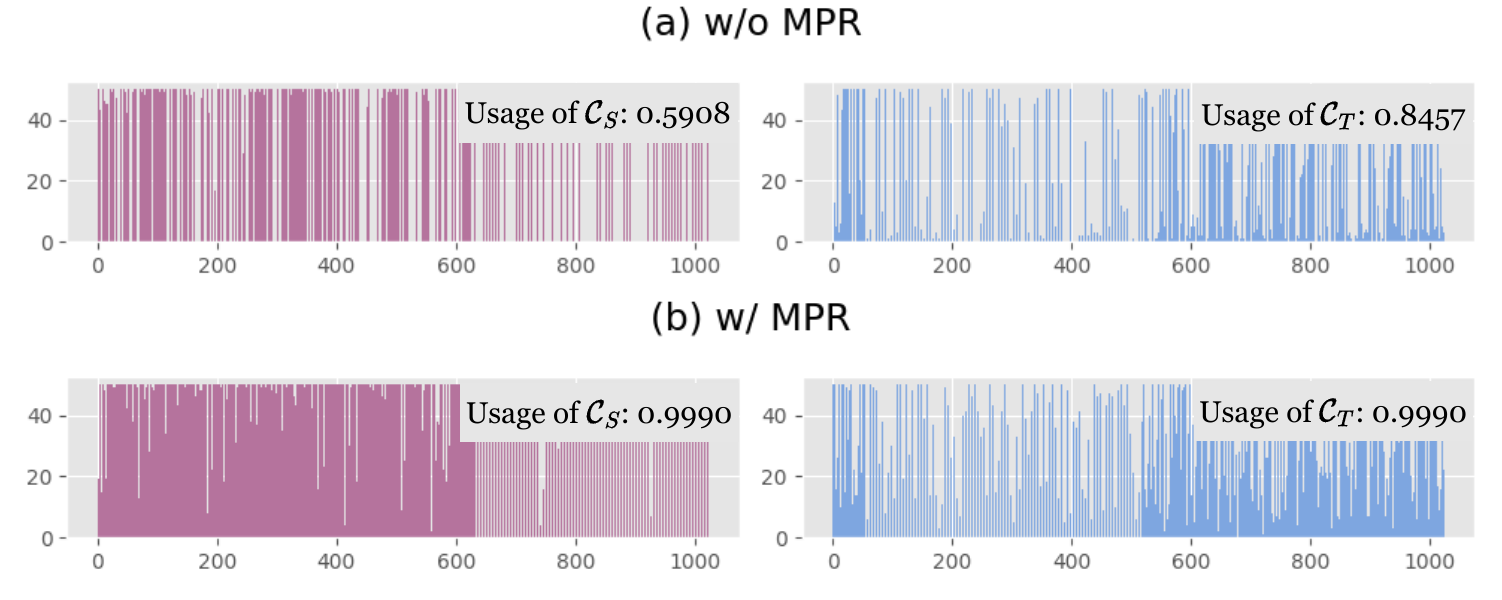}
    \caption{Comparison of \colorbox[HTML]{B5739D}{spatial} and \colorbox[HTML]{7EA6E0}{temporal} codebooks' utilization when applying different regualizations. }
    \label{fig:codebook}
\end{figure}

\begin{table*}[ht]
    \centering
    \tabcolsep=3pt
    \begin{subtable}[t]{0.32\linewidth}
    \centering
        \begin{tabular}{cccc}
            Arch & Objective & SSIM$\uparrow$ & FVD$\downarrow$ \\ \toprule
            ViT-S& DINOv2   &\cellcolor[HTML]{E6E6E6}\textbf{0.9054} &\cellcolor[HTML]{E6E6E6}49.11 \\
            ViT-B& DINOv2   &0.9050 &\textbf{49.08} \\
            ViT-B& CLIP     &0.8935 &66.86 \\
        \end{tabular}
        \caption{\textbf{Discriminator feature networks.}}
    \end{subtable}
    \hspace{0.1cm}
    \begin{subtable}[t]{0.32\linewidth}
    \centering
        \begin{tabular}{lcc}
            Fusion $\oplus$ & SSIM$\uparrow$ & FVD$\downarrow$ \\ \toprule
            Add &\cellcolor[HTML]{E6E6E6}\textbf{0.9054} &\cellcolor[HTML]{E6E6E6}\textbf{49.11} \\
            Conv &0.8799 &97.67 \\
            3DFFT~\cite{lu2024freelong} &0.8741 &118.7 \\
        \end{tabular}
        \caption{\textbf{Fusion operation $\oplus$} between quantized spatial and temporal latents.}
    \end{subtable} 
    \hspace{0.1cm}
    \begin{subtable}[t]{0.32\linewidth}
    \centering
        \begin{tabular}{ccccc}
            $\mathcal{L}_{base}$ & $\mathcal{L}_{adv}$ & $\mathcal{L}_{KL}$ & SSIM$\uparrow$ & FVD$\downarrow$ \\ \toprule
            $\usym{2713}$ &$\usym{2717}$ &$\usym{2717}$ &0.8887 &66.82 \\
            $\usym{2713}$ &$\usym{2713}$ &$\usym{2717}$ &0.8914 &65.90 \\
            $\usym{2713}$ &$\usym{2713}$ &$\usym{2713}$ &\cellcolor[HTML]{E6E6E6}\textbf{0.9054} &\cellcolor[HTML]{E6E6E6}\textbf{49.11} \\
        \end{tabular}
        \caption{\textbf{Loss terms} for Stage \Rmnum{1}. 
        }
    \end{subtable}
    \\ 
    \begin{subtable}[t]{0.32\linewidth}
    \centering
        \begin{tabular}{lcc}
            Regularization & SSIM$\uparrow$ & FVD$\downarrow$ \\ \toprule
            None &0.8914 &65.90 \\
            PDR~\cite{zhang2023regularized} &0.9015 &56.46 \\
            CVQ~\cite{zheng2023online} &0.4667 &940.1 \\
            MPR(Proposed) &\cellcolor[HTML]{E6E6E6}\textbf{0.9054} &\cellcolor[HTML]{E6E6E6}\textbf{49.11} \\
        \end{tabular}
        \caption{\textbf{Codebook regularization.}}
    \end{subtable}
    \hspace{0.1cm}
    \begin{subtable}[t]{0.32\linewidth}
    \centering
        \begin{tabular}{ccccc}
            $N_S$ & $N_T$ &$D$ & SSIM$\uparrow$ & FVD$\downarrow$  \\ \toprule
            512 &512 &256 &0.8919 &69.28 \\ 
            1024 &1024 &256 &\cellcolor[HTML]{E6E6E6}\textbf{0.9054} &\cellcolor[HTML]{E6E6E6}\textbf{49.11} \\
            2048 &2048 &256 &0.8934 &66.29 \\
            1024 &1024 &64 &0.8871 &71.31 \\
        \end{tabular}
        \caption{\textbf{Codebook size and dimension.}}
    \end{subtable}
    \hspace{0.1cm}
    \begin{subtable}[t]{0.32\linewidth}
    \centering
        \begin{tabular}{cccccc}
            $\mathbb{A}$ &$\mathbb{B}$ &$\mathbb{C}$ & \#Video & SSIM$\uparrow$ & FVD$\downarrow$\\ \toprule
            $\usym{2717}$ &$\usym{2717}$ &$\usym{2717}$ &15967 &0.8891 &71.67\\
            $\usym{2713}$ &$\usym{2717}$ &$\usym{2717}$ &15967 &0.8918 &67.62\\
            $\usym{2713}$ &$\usym{2713}$ &$\usym{2717}$ &12216 &0.8995 &61.81\\
            $\usym{2713}$ &$\usym{2713}$ &$\usym{2713}$ &11057 &\cellcolor[HTML]{E6E6E6}\textbf{0.9054} &\cellcolor[HTML]{E6E6E6}\textbf{49.11} \\
        \end{tabular}
        \caption{\textbf{Data manipulation.} }
    \end{subtable}
    \caption{\textbf{Ablation study.} 
    We report SSIM and FVD results of different variants for reconstruction on the VFHQ-test dataset. 
    The training length is 250,000 iterations. 
    Default settings are marked as \colorbox[HTML]{E6E6E6}{gray}.}
    \label{tab:ablation}
\end{table*}

We conducted an ablation study to evaluate several design choices, which we discuss in detail below.

\noindent \textbf{Discriminator feature networks.} 
(Table~\ref{tab:ablation}(a)). 
The work in~\cite{sauer2023adversarial} indicates that applying pre-trained feature networks, such as CLIP and DINO, as the discriminator can boost the performance of diffusion distillation. 
We adopt a pre-trained network to serve as a more powerful discriminator to reduce artifacts and stabilize training. 
The results show that DINO's objective is more suitable than CLIP's and the scale of ViT has little effect on the performance.

\noindent \textbf{Fusion operation.} 
(Table~\ref{tab:ablation}(b)). 
For the fusion operation $\oplus$ between quantized spatial latent $z_{q,S}$ and temporal latent $z_{q,T}$, we have tried various fusion methods, such as direct addition, convolution projection, and 3DFFT fusion~\cite{lu2024freelong}. 
Convolution projection means concatenating two latents along the channel dimension and using 3D convolution to reduce it to normal size. 
3DFFT represents the fusion of latents by processing two variables separately by 3DFFT with a pre-defined pass threshold. 
Among these methods, direct addition yields the best performance. 

\noindent \textbf{Loss terms.} 
(Table~\ref{tab:ablation}(c)). 
The results in the table show the contribution of the discriminator and proposed marginal prior regularization in the training of Stage \Rmnum{1}. 

\noindent \textbf{Codebook regularization.} 
(Table~\ref{tab:ablation}(d)). 
We report the reconstruction performance in Stage \Rmnum{1} under different regularization. 
PDR and MPR contribute to the learning of codebooks and backbones compared with the vanilla version. 
CVQ failed when jointly optimizing multiple codebooks, leading to model collapse. 
Figure~\ref{fig:codebook}(a) and (b) display the comparison of codebook utilization when reconstructing the videos in the VFHQ-Test dataset. 
Applying the proposed MPR can effectively alleviate the codebook collapse problem and improve the utilization of codebooks. 

\noindent \textbf{Codebook size and dimension.} 
(Table~\ref{tab:ablation}(e)). 
Experiments show that a moderate codebook size and latent space dimension improve performance.

\noindent \textbf{Data manipulation.} 
(Table~\ref{tab:ablation}(f)). 
We preprocess training videos in three aspects, comprising face proportion ($\mathbb{A}$), face orientation ($\mathbb{B}$), and the presence of text ($\mathbb{C}$). 
From the results, we can learn that compared to the quantity, the quality of face videos is more helpful for the learning of 3D-VQGAN backbone and spatial-temporal codebooks. 

\section{Conclusion}

In this work, we propose a two-stage video face enhancement framework, which can jointly solve BFVR and de-flickering tasks. 
This framework develops upon a 3D-VQGAN backbone and a more powerful discriminator for efficient video compression and spatial-temporal codebooks for effective video quantization. 
We also design a marginal prior regularization to mitigate the codebook collapse problem. 
Experiments demonstrate that our method achieves promising restoration and de-flickering performance with enhanced inference efficiency. 
\newpage
{
    \small
    \bibliographystyle{ieeenat_fullname}
    \bibliography{main}

\begin{thebibliography}{58}
\providecommand{\natexlab}[1]{#1}
\providecommand{\url}[1]{\texttt{#1}}
\expandafter\ifx\csname urlstyle\endcsname\relax
  \providecommand{\doi}[1]{doi: #1}\else
  \providecommand{\doi}{doi: \begingroup \urlstyle{rm}\Url}\fi

\bibitem[Baevski et~al.(2019)Baevski, Schneider, and Auli]{baevski2019vq}
Alexei Baevski, Steffen Schneider, and Michael Auli.
\newblock vq-wav2vec: Self-supervised learning of discrete speech representations.
\newblock \emph{arXiv preprint arXiv:1910.05453}, 2019.

\bibitem[Bonneel et~al.(2015)Bonneel, Tompkin, Sunkavalli, Sun, Paris, and Pfister]{bonneel2015blind}
Nicolas Bonneel, James Tompkin, Kalyan Sunkavalli, Deqing Sun, Sylvain Paris, and Hanspeter Pfister.
\newblock Blind video temporal consistency.
\newblock \emph{ACM Transactions on Graphics (TOG)}, 34\penalty0 (6):\penalty0 1--9, 2015.

\bibitem[Chan et~al.(2021)Chan, Wang, Xu, Gu, and Loy]{chan2021glean}
Kelvin~CK Chan, Xintao Wang, Xiangyu Xu, Jinwei Gu, and Chen~Change Loy.
\newblock Glean: Generative latent bank for large-factor image super-resolution.
\newblock In \emph{Proceedings of the IEEE/CVF conference on computer vision and pattern recognition}, pages 14245--14254, 2021.

\bibitem[Chan et~al.(2022{\natexlab{a}})Chan, Zhou, Xu, and Loy]{chan2022basicvsr++}
Kelvin~CK Chan, Shangchen Zhou, Xiangyu Xu, and Chen~Change Loy.
\newblock Basicvsr++: Improving video super-resolution with enhanced propagation and alignment.
\newblock In \emph{Proceedings of the IEEE/CVF conference on computer vision and pattern recognition}, pages 5972--5981, 2022{\natexlab{a}}.

\bibitem[Chan et~al.(2022{\natexlab{b}})Chan, Zhou, Xu, and Loy]{chan2022investigating}
Kelvin~CK Chan, Shangchen Zhou, Xiangyu Xu, and Chen~Change Loy.
\newblock Investigating tradeoffs in real-world video super-resolution.
\newblock In \emph{Proceedings of the IEEE/CVF Conference on Computer Vision and Pattern Recognition}, pages 5962--5971, 2022{\natexlab{b}}.

\bibitem[Chen et~al.(2021)Chen, Li, Yang, Lin, Zhang, and Wong]{chen2021progressive}
Chaofeng Chen, Xiaoming Li, Lingbo Yang, Xianhui Lin, Lei Zhang, and Kwan-Yee~K Wong.
\newblock Progressive semantic-aware style transformation for blind face restoration.
\newblock In \emph{Proceedings of the IEEE/CVF conference on computer vision and pattern recognition}, pages 11896--11905, 2021.

\bibitem[Chen et~al.(2018)Chen, Tai, Liu, Shen, and Yang]{chen2018fsrnet}
Yu Chen, Ying Tai, Xiaoming Liu, Chunhua Shen, and Jian Yang.
\newblock Fsrnet: End-to-end learning face super-resolution with facial priors.
\newblock In \emph{Proceedings of the IEEE conference on computer vision and pattern recognition}, pages 2492--2501, 2018.

\bibitem[Delon and Desolneux(2010)]{delon2010stabilization}
Julie Delon and Agnes Desolneux.
\newblock Stabilization of flicker-like effects in image sequences through local contrast correction.
\newblock \emph{SIAM Journal on Imaging Sciences}, 3\penalty0 (4):\penalty0 703--734, 2010.

\bibitem[Duan et~al.(2023)Duan, Wang, Chen, Qian, Huang, and Jin]{duan2023fastblend}
Zhongjie Duan, Chengyu Wang, Cen Chen, Weining Qian, Jun Huang, and Mingyi Jin.
\newblock Fastblend: a powerful model-free toolkit making video stylization easier.
\newblock \emph{arXiv preprint arXiv:2311.09265}, 2023.

\bibitem[Duan et~al.(2024)Duan, Wang, Chen, Qian, and Huang]{duan2024diffutoon}
Zhongjie Duan, Chengyu Wang, Cen Chen, Weining Qian, and Jun Huang.
\newblock Diffutoon: High-resolution editable toon shading via diffusion models.
\newblock \emph{arXiv preprint arXiv:2401.16224}, 2024.

\bibitem[Esser et~al.(2021)Esser, Rombach, and Ommer]{esser2021taming}
Patrick Esser, Robin Rombach, and Bjorn Ommer.
\newblock Taming transformers for high-resolution image synthesis.
\newblock In \emph{Proceedings of the IEEE/CVF conference on computer vision and pattern recognition}, pages 12873--12883, 2021.

\bibitem[Feng et~al.(2024)Feng, Li, and Loy]{feng2024kalman}
Ruicheng Feng, Chongyi Li, and Chen~Change Loy.
\newblock Kalman-inspired feature propagation for video face super-resolution.
\newblock \emph{arXiv preprint arXiv:2408.05205}, 2024.

\bibitem[Fulari et~al.(2024)Fulari, Mulleti, and Rajwade]{fulari2024unsupervised}
Anuj Fulari, Satish Mulleti, and Ajit Rajwade.
\newblock Unsupervised model-based learning for simultaneous video deflickering and deblotching.
\newblock In \emph{Proceedings of the IEEE/CVF Winter Conference on Applications of Computer Vision}, pages 4117--4125, 2024.

\bibitem[Gu et~al.(2022)Gu, Wang, Xie, Dong, Li, Shan, and Cheng]{gu2022vqfr}
Yuchao Gu, Xintao Wang, Liangbin Xie, Chao Dong, Gen Li, Ying Shan, and Ming-Ming Cheng.
\newblock Vqfr: Blind face restoration with vector-quantized dictionary and parallel decoder.
\newblock In \emph{European Conference on Computer Vision}, pages 126--143. Springer, 2022.

\bibitem[Hu et~al.(2020)Hu, Ren, LaMaster, Cao, Li, Li, Menze, and Liu]{hu2020face}
Xiaobin Hu, Wenqi Ren, John LaMaster, Xiaochun Cao, Xiaoming Li, Zechao Li, Bjoern Menze, and Wei Liu.
\newblock Face super-resolution guided by 3d facial priors.
\newblock In \emph{Computer Vision--ECCV 2020: 16th European Conference, Glasgow, UK, August 23--28, 2020, Proceedings, Part IV 16}, pages 763--780. Springer, 2020.

\bibitem[Huh et~al.(2023)Huh, Cheung, Agrawal, and Isola]{huh2023straightening}
Minyoung Huh, Brian Cheung, Pulkit Agrawal, and Phillip Isola.
\newblock Straightening out the straight-through estimator: Overcoming optimization challenges in vector quantized networks.
\newblock In \emph{International Conference on Machine Learning}, pages 14096--14113. PMLR, 2023.

\bibitem[Johnson et~al.(2016)Johnson, Alahi, and Fei-Fei]{johnson2016perceptual}
Justin Johnson, Alexandre Alahi, and Li Fei-Fei.
\newblock Perceptual losses for real-time style transfer and super-resolution.
\newblock In \emph{Computer Vision--ECCV 2016: 14th European Conference, Amsterdam, The Netherlands, October 11-14, 2016, Proceedings, Part II 14}, pages 694--711. Springer, 2016.

\bibitem[Kanj et~al.(2017)Kanj, Talbot, and Luparello]{kanj2017flicker}
Ali Kanj, Hugues Talbot, and Raoul~Rodriguez Luparello.
\newblock Flicker removal and superpixel-based motion tracking for high speed videos.
\newblock In \emph{2017 IEEE international conference on image processing (ICIP)}, pages 245--249. IEEE, 2017.

\bibitem[Kappeler et~al.(2016)Kappeler, Yoo, Dai, and Katsaggelos]{kappeler2016video}
Armin Kappeler, Seunghwan Yoo, Qiqin Dai, and Aggelos~K Katsaggelos.
\newblock Video super-resolution with convolutional neural networks.
\newblock \emph{IEEE transactions on computational imaging}, 2\penalty0 (2):\penalty0 109--122, 2016.

\bibitem[Karras et~al.(2019)Karras, Laine, and Aila]{karras2019style}
Tero Karras, Samuli Laine, and Timo Aila.
\newblock A style-based generator architecture for generative adversarial networks.
\newblock In \emph{Proceedings of the IEEE/CVF conference on computer vision and pattern recognition}, pages 4401--4410, 2019.

\bibitem[Karras et~al.(2020)Karras, Laine, Aittala, Hellsten, Lehtinen, and Aila]{karras2020analyzing}
Tero Karras, Samuli Laine, Miika Aittala, Janne Hellsten, Jaakko Lehtinen, and Timo Aila.
\newblock Analyzing and improving the image quality of stylegan.
\newblock In \emph{Proceedings of the IEEE/CVF conference on computer vision and pattern recognition}, pages 8110--8119, 2020.

\bibitem[Kingma(2014)]{kingma2014adam}
Diederik~P Kingma.
\newblock Adam: A method for stochastic optimization.
\newblock \emph{arXiv preprint arXiv:1412.6980}, 2014.

\bibitem[Lab and etc.(2024)]{pku_yuan_lab_and_tuzhan_ai_etc_2024_10948109}
PKU-Yuan Lab and Tuzhan~AI etc.
\newblock Open-sora-plan, 2024.

\bibitem[{\L}a{\'n}cucki et~al.(2020){\L}a{\'n}cucki, Chorowski, Sanchez, Marxer, Chen, Dolfing, Khurana, Alum{\"a}e, and Laurent]{lancucki2020robust}
Adrian {\L}a{\'n}cucki, Jan Chorowski, Guillaume Sanchez, Ricard Marxer, Nanxin Chen, Hans~JGA Dolfing, Sameer Khurana, Tanel Alum{\"a}e, and Antoine Laurent.
\newblock Robust training of vector quantized bottleneck models.
\newblock In \emph{2020 International Joint Conference on Neural Networks (IJCNN)}, pages 1--7. IEEE, 2020.

\bibitem[Lei et~al.(2020)Lei, Xing, and Chen]{lei2020blind}
Chenyang Lei, Yazhou Xing, and Qifeng Chen.
\newblock Blind video temporal consistency via deep video prior.
\newblock \emph{Advances in Neural Information Processing Systems}, 33:\penalty0 1083--1093, 2020.

\bibitem[Lei et~al.(2023)Lei, Ren, Zhang, and Chen]{lei2023blind}
Chenyang Lei, Xuanchi Ren, Zhaoxiang Zhang, and Qifeng Chen.
\newblock Blind video deflickering by neural filtering with a flawed atlas.
\newblock In \emph{Proceedings of the IEEE/CVF Conference on Computer Vision and Pattern Recognition}, pages 10439--10448, 2023.

\bibitem[Li et~al.(2018)Li, Liu, Ye, Zuo, Lin, and Yang]{li2018learning}
Xiaoming Li, Ming Liu, Yuting Ye, Wangmeng Zuo, Liang Lin, and Ruigang Yang.
\newblock Learning warped guidance for blind face restoration.
\newblock In \emph{Proceedings of the European conference on computer vision (ECCV)}, pages 272--289, 2018.

\bibitem[Liang et~al.(2024)Liang, Cao, Fan, Zhang, Ranjan, Li, Timofte, and Van~Gool]{liang2024vrt}
Jingyun Liang, Jiezhang Cao, Yuchen Fan, Kai Zhang, Rakesh Ranjan, Yawei Li, Radu Timofte, and Luc Van~Gool.
\newblock Vrt: A video restoration transformer.
\newblock \emph{IEEE Transactions on Image Processing}, 2024.

\bibitem[Lin et~al.(2023)Lin, He, Chen, Lyu, Dai, Yu, Ouyang, Qiao, and Dong]{lin2023diffbir}
Xinqi Lin, Jingwen He, Ziyan Chen, Zhaoyang Lyu, Bo Dai, Fanghua Yu, Wanli Ouyang, Yu Qiao, and Chao Dong.
\newblock Diffbir: Towards blind image restoration with generative diffusion prior.
\newblock \emph{arXiv preprint arXiv:2308.15070}, 2023.

\bibitem[Liu et~al.(2024)Liu, Cun, Liu, Wang, Zhang, Chen, Liu, Zeng, Chan, and Shan]{liu2024evalcrafter}
Yaofang Liu, Xiaodong Cun, Xuebo Liu, Xintao Wang, Yong Zhang, Haoxin Chen, Yang Liu, Tieyong Zeng, Raymond Chan, and Ying Shan.
\newblock Evalcrafter: Benchmarking and evaluating large video generation models.
\newblock In \emph{Proceedings of the IEEE/CVF Conference on Computer Vision and Pattern Recognition}, pages 22139--22149, 2024.

\bibitem[Lu et~al.(2024)Lu, Liang, Zhu, and Yang]{lu2024freelong}
Yu Lu, Yuanzhi Liang, Linchao Zhu, and Yi Yang.
\newblock Freelong: Training-free long video generation with spectralblend temporal attention.
\newblock \emph{arXiv preprint arXiv:2407.19918}, 2024.

\bibitem[OpenAI(2024)]{openai2024sora}
OpenAI.
\newblock Sora: Creating video from text., 2024.
\newblock \url{https://openai.com/sora/}.

\bibitem[Oquab et~al.(2023)Oquab, Darcet, Moutakanni, Vo, Szafraniec, Khalidov, Fernandez, Haziza, Massa, El-Nouby, et~al.]{oquab2023dinov2}
Maxime Oquab, Timoth{\'e}e Darcet, Th{\'e}o Moutakanni, Huy Vo, Marc Szafraniec, Vasil Khalidov, Pierre Fernandez, Daniel Haziza, Francisco Massa, Alaaeldin El-Nouby, et~al.
\newblock Dinov2: Learning robust visual features without supervision.
\newblock \emph{arXiv preprint arXiv:2304.07193}, 2023.

\bibitem[Razavi et~al.(2019)Razavi, Van~den Oord, and Vinyals]{razavi2019generating}
Ali Razavi, Aaron Van~den Oord, and Oriol Vinyals.
\newblock Generating diverse high-fidelity images with vq-vae-2.
\newblock \emph{Advances in neural information processing systems}, 32, 2019.

\bibitem[Rombach et~al.(2022)Rombach, Blattmann, Lorenz, Esser, and Ommer]{rombach2022high}
Robin Rombach, Andreas Blattmann, Dominik Lorenz, Patrick Esser, and Bj{\"o}rn Ommer.
\newblock High-resolution image synthesis with latent diffusion models.
\newblock In \emph{Proceedings of the IEEE/CVF conference on computer vision and pattern recognition}, pages 10684--10695, 2022.

\bibitem[Sauer et~al.(2023)Sauer, Lorenz, Blattmann, and Rombach]{sauer2023adversarial}
Axel Sauer, Dominik Lorenz, Andreas Blattmann, and Robin Rombach.
\newblock Adversarial diffusion distillation.
\newblock \emph{arXiv preprint arXiv:2311.17042}, 2023.

\bibitem[Sauer et~al.(2024)Sauer, Boesel, Dockhorn, Blattmann, Esser, and Rombach]{sauer2024fast}
Axel Sauer, Frederic Boesel, Tim Dockhorn, Andreas Blattmann, Patrick Esser, and Robin Rombach.
\newblock Fast high-resolution image synthesis with latent adversarial diffusion distillation.
\newblock \emph{arXiv preprint arXiv:2403.12015}, 2024.

\bibitem[Shi et~al.(2016)Shi, Caballero, Husz{\'a}r, Totz, Aitken, Bishop, Rueckert, and Wang]{shi2016real}
Wenzhe Shi, Jose Caballero, Ferenc Husz{\'a}r, Johannes Totz, Andrew~P Aitken, Rob Bishop, Daniel Rueckert, and Zehan Wang.
\newblock Real-time single image and video super-resolution using an efficient sub-pixel convolutional neural network.
\newblock In \emph{Proceedings of the IEEE conference on computer vision and pattern recognition}, pages 1874--1883, 2016.

\bibitem[Simonyan and Zisserman(2015)]{simonyan2015very}
K Simonyan and A Zisserman.
\newblock Very deep convolutional networks for large-scale image recognition.
\newblock In \emph{3rd International Conference on Learning Representations (ICLR 2015)}. Computational and Biological Learning Society, 2015.

\bibitem[Takida et~al.(2022)Takida, Shibuya, Liao, Lai, Ohmura, Uesaka, Murata, Takahashi, Kumakura, and Mitsufuji]{takida2022sq}
Yuhta Takida, Takashi Shibuya, WeiHsiang Liao, Chieh-Hsin Lai, Junki Ohmura, Toshimitsu Uesaka, Naoki Murata, Shusuke Takahashi, Toshiyuki Kumakura, and Yuki Mitsufuji.
\newblock Sq-vae: Variational bayes on discrete representation with self-annealed stochastic quantization.
\newblock \emph{arXiv preprint arXiv:2205.07547}, 2022.

\bibitem[Tan et~al.(2024)Tan, Park, Zhang, Wang, Zhang, Kong, Dai, Liu, and Luo]{tan2024blind}
Jingfan Tan, Hyunhee Park, Ying Zhang, Tao Wang, Kaihao Zhang, Xiangyu Kong, Pengwen Dai, Zikun Liu, and Wenhan Luo.
\newblock Blind face video restoration with temporal consistent generative prior and degradation-aware prompt.
\newblock In \emph{ACM Multimedia 2024}, 2024.

\bibitem[Teed and Deng(2020)]{teed2020raft}
Zachary Teed and Jia Deng.
\newblock Raft: Recurrent all-pairs field transforms for optical flow.
\newblock In \emph{Computer Vision--ECCV 2020: 16th European Conference, Glasgow, UK, August 23--28, 2020, Proceedings, Part II 16}, pages 402--419. Springer, 2020.

\bibitem[Van Den~Oord et~al.(2017)Van Den~Oord, Vinyals, et~al.]{van2017neural}
Aaron Van Den~Oord, Oriol Vinyals, et~al.
\newblock Neural discrete representation learning.
\newblock \emph{Advances in neural information processing systems}, 30, 2017.

\bibitem[Wang et~al.(2019)Wang, Chan, Yu, Dong, and Change~Loy]{wang2019edvr}
Xintao Wang, Kelvin~CK Chan, Ke Yu, Chao Dong, and Chen Change~Loy.
\newblock Edvr: Video restoration with enhanced deformable convolutional networks.
\newblock In \emph{Proceedings of the IEEE/CVF conference on computer vision and pattern recognition workshops}, pages 0--0, 2019.

\bibitem[Wang et~al.(2021{\natexlab{a}})Wang, Li, Zhang, and Shan]{wang2021towards}
Xintao Wang, Yu Li, Honglun Zhang, and Ying Shan.
\newblock Towards real-world blind face restoration with generative facial prior.
\newblock In \emph{Proceedings of the IEEE/CVF conference on computer vision and pattern recognition}, pages 9168--9178, 2021{\natexlab{a}}.

\bibitem[Wang et~al.(2021{\natexlab{b}})Wang, Xie, Dong, and Shan]{wang2021real}
Xintao Wang, Liangbin Xie, Chao Dong, and Ying Shan.
\newblock Real-esrgan: Training real-world blind super-resolution with pure synthetic data.
\newblock In \emph{Proceedings of the IEEE/CVF international conference on computer vision}, pages 1905--1914, 2021{\natexlab{b}}.

\bibitem[Wang et~al.(2022)Wang, Zhang, Chen, Wang, and Luo]{wang2022restoreformer}
Zhouxia Wang, Jiawei Zhang, Runjian Chen, Wenping Wang, and Ping Luo.
\newblock Restoreformer: High-quality blind face restoration from undegraded key-value pairs.
\newblock In \emph{Proceedings of the IEEE/CVF conference on computer vision and pattern recognition}, pages 17512--17521, 2022.

\bibitem[Xie et~al.(2022)Xie, Wang, Zhang, Dong, and Shan]{xie2022vfhq}
Liangbin Xie, Xintao Wang, Honglun Zhang, Chao Dong, and Ying Shan.
\newblock Vfhq: A high-quality dataset and benchmark for video face super-resolution.
\newblock In \emph{Proceedings of the IEEE/CVF Conference on Computer Vision and Pattern Recognition}, pages 657--666, 2022.

\bibitem[Xu et~al.(2024)Xu, Xu, He, Yu, and Li]{xu2024beyond}
Kepeng Xu, Li Xu, Gang He, Wenxin Yu, and Yunsong Li.
\newblock Beyond alignment: Blind video face restoration via parsing-guided temporal-coherent transformer.
\newblock \emph{arXiv preprint arXiv:2404.13640}, 2024.

\bibitem[Yang et~al.(2021)Yang, Ren, Xie, and Zhang]{yang2021gan}
Tao Yang, Peiran Ren, Xuansong Xie, and Lei Zhang.
\newblock Gan prior embedded network for blind face restoration in the wild.
\newblock In \emph{Proceedings of the IEEE/CVF conference on computer vision and pattern recognition}, pages 672--681, 2021.

\bibitem[Yatziv and Sapiro(2006)]{yatziv2006fast}
Liron Yatziv and Guillermo Sapiro.
\newblock Fast image and video colorization using chrominance blending.
\newblock \emph{IEEE transactions on image processing}, 15\penalty0 (5):\penalty0 1120--1129, 2006.

\bibitem[Zeghidour et~al.(2021)Zeghidour, Luebs, Omran, Skoglund, and Tagliasacchi]{zeghidour2021soundstream}
Neil Zeghidour, Alejandro Luebs, Ahmed Omran, Jan Skoglund, and Marco Tagliasacchi.
\newblock Soundstream: An end-to-end neural audio codec.
\newblock \emph{IEEE/ACM Transactions on Audio, Speech, and Language Processing}, 30:\penalty0 495--507, 2021.

\bibitem[Zhang et~al.(2023)Zhang, Zhan, Theobalt, and Lu]{zhang2023regularized}
Jiahui Zhang, Fangneng Zhan, Christian Theobalt, and Shijian Lu.
\newblock Regularized vector quantization for tokenized image synthesis.
\newblock In \emph{Proceedings of the IEEE/CVF Conference on Computer Vision and Pattern Recognition}, pages 18467--18476, 2023.

\bibitem[Zhang et~al.(2018)Zhang, Isola, Efros, Shechtman, and Wang]{zhang2018unreasonable}
Richard Zhang, Phillip Isola, Alexei~A Efros, Eli Shechtman, and Oliver Wang.
\newblock The unreasonable effectiveness of deep features as a perceptual metric.
\newblock In \emph{Proceedings of the IEEE conference on computer vision and pattern recognition}, pages 586--595, 2018.

\bibitem[Zhao et~al.(2023)Zhao, Po, Liu, Wang, Yu, Xian, Zhang, and Liu]{zhao2023svcnet}
Yuzhi Zhao, Lai-Man Po, Kangcheng Liu, Xuehui Wang, Wing-Yin Yu, Pengfei Xian, Yujia Zhang, and Mengyang Liu.
\newblock Svcnet: Scribble-based video colorization network with temporal aggregation.
\newblock \emph{IEEE Transactions on Image Processing}, 2023.

\bibitem[Zheng and Vedaldi(2023)]{zheng2023online}
Chuanxia Zheng and Andrea Vedaldi.
\newblock Online clustered codebook.
\newblock In \emph{Proceedings of the IEEE/CVF International Conference on Computer Vision}, pages 22798--22807, 2023.

\bibitem[Zheng et~al.(2022)Zheng, Vuong, Cai, and Phung]{zheng2022movq}
Chuanxia Zheng, Tung-Long Vuong, Jianfei Cai, and Dinh Phung.
\newblock Movq: Modulating quantized vectors for high-fidelity image generation.
\newblock \emph{Advances in Neural Information Processing Systems}, 35:\penalty0 23412--23425, 2022.

\bibitem[Zhou et~al.(2022)Zhou, Chan, Li, and Loy]{zhou2022towards}
Shangchen Zhou, Kelvin Chan, Chongyi Li, and Chen~Change Loy.
\newblock Towards robust blind face restoration with codebook lookup transformer.
\newblock \emph{Advances in Neural Information Processing Systems}, 35:\penalty0 30599--30611, 2022.

\end{thebibliography}
}

\clearpage
\setcounter{page}{1}
\maketitlesupplementary


\vspace{-0.5cm}
\section{Dataset Analysis}

\begin{figure}[t]
    \centering
    \includegraphics[width=0.9\linewidth]{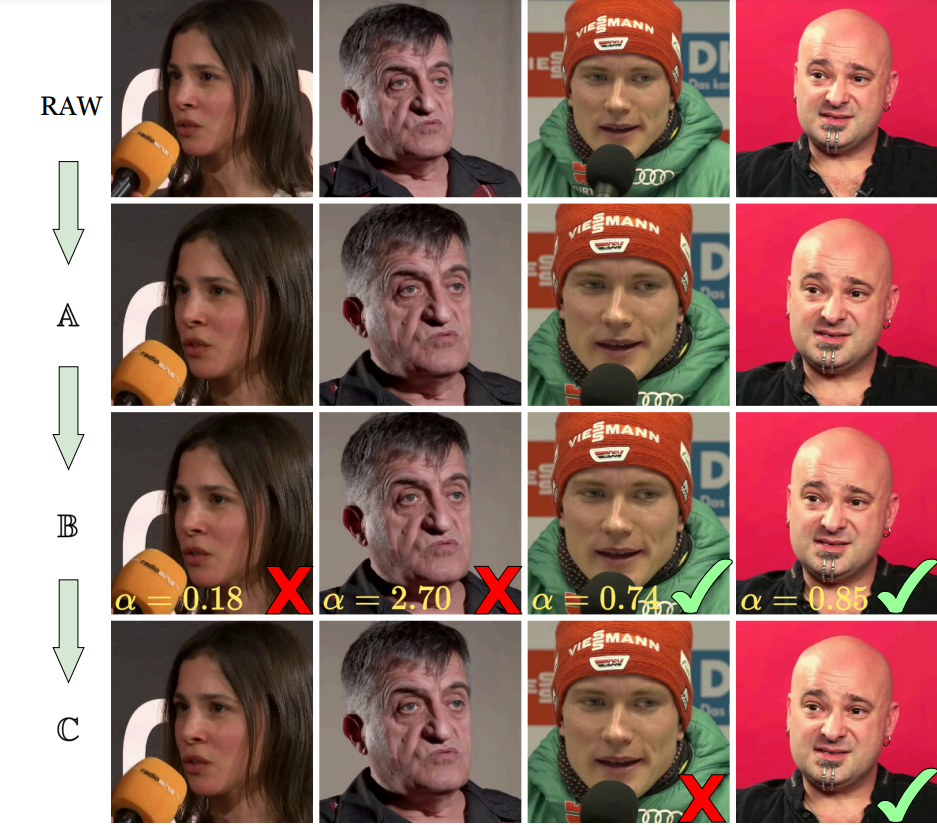}
    \caption{Visualization of the training data processing, including face proportion ($\mathbb{A}$), face orientation ($\mathbb{B}$), and text presence ($\mathbb{C}$).}
    \label{fig:dataset_illu}
\end{figure}

As illustrated in Figure~\ref{fig:dataset_illu}, we preprocess the training videos in three aspects successively, including \textbf{face proportion ($\mathbb{A}$)}, \textbf{face orientation ($\mathbb{B}$)}, and \textbf{the presence of text ($\mathbb{C}$)}. 
In step $\mathbb{A}$ (second row in Figure~\ref{fig:dataset_illu}), we crop the raw videos to increase the face proportion. 
Subsequently, in step $\mathbb{B}$ (third row), we compute the side ratio $\alpha$ and exclude videos with prominent side faces (e.g., first and second column), which may hinder the learning of high-quality frontal portrait features. 
Finally, in step $\mathbb{C}$ (fourth row), we employ PaddleOCR to filter out videos containing text elements (e.g., third column), which exist in backgrounds or clothing.

\subsection{Face orientation}
We filter side faces in videos using facial structural analysis. 
For each frame, we first convert it into a grayscale image and then utilize Dlib to generate facial landmarks. 
The outer corners of the left and right eyes are defined as coordinates $(x_1, y_1)$ and $(x_2, y_2)$, respectively, while the nose tip is represented by $(x_0, y_0)$. 
\begin{itemize}
    \item Initially, we conduct a preliminary screening based on the relative positions of the eyes and nose. Specifically, if $x_1 > x_0$ or $x_2 < x_0$, the face is categorized as a side face. 
    \item Subsequently, a manually defined threshold is employed to further filter based on the side ratio $\alpha=\frac{dis_l}{dis_r}=\frac{|x_1-x_0|}{|x_2-x_0|}$. A face is categorized as a side face if $\alpha<0.4$ or $\alpha>2.5$. 
\end{itemize}

\subsection{Motion intensity}
\begin{figure}[t]
    \centering
    \includegraphics[width=1.0\linewidth]{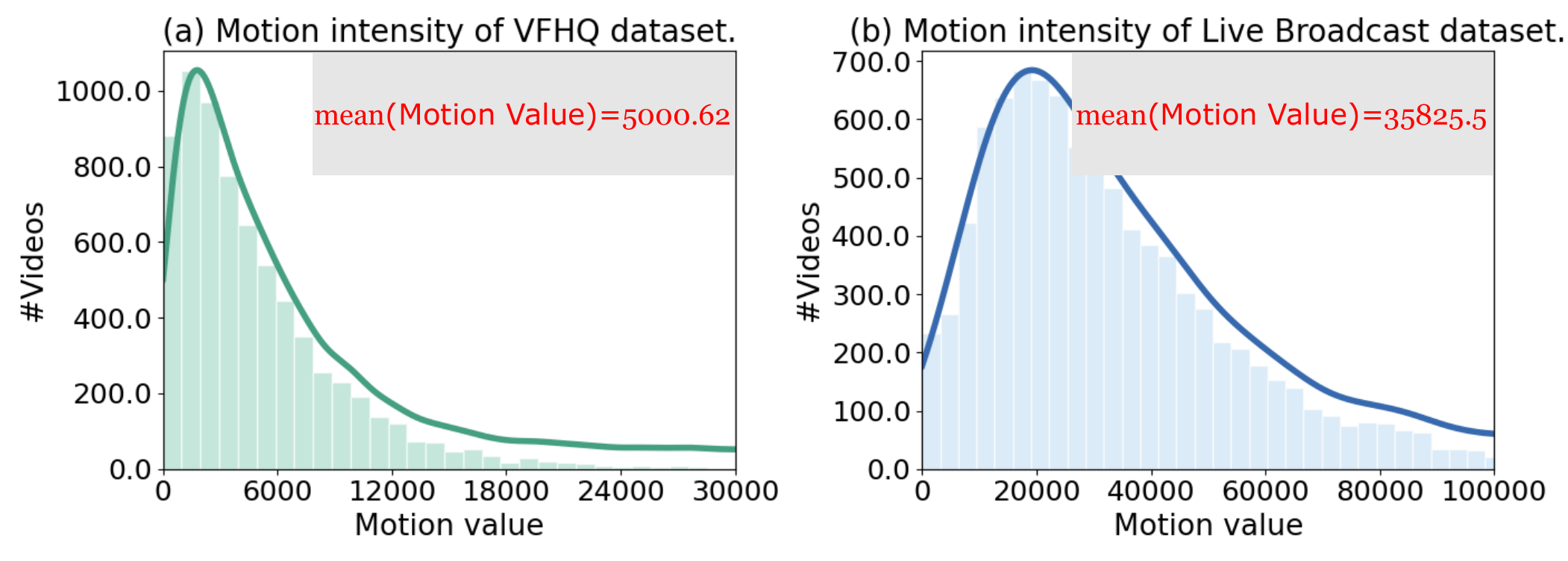}
    \caption{Visualization of motion intensity of (a) 
 VFHQ and (b) Live Broadcast dataset.}
    \label{fig:motion_comp}
\end{figure}

\vspace{-0.2cm}
In our experiments, we found that the motion intensity of training videos significantly influences the learning of the proposed spatial-temporal codebooks. 
Videos with high motion intensity provide richer temporal residuals, thereby enhancing the temporal codebook's expressiveness. 
In this study, motion intensity is calculated as the sum of non-zero pixel points after binarizing the difference between adjacent grayscale frames. 
We illustrate the motion intensity of the VFHQ dataset and our collected LiveBroadcast dataset in Figure~\ref{fig:motion_comp}. 
Compared to the LiveBroadcast dataset, VFHQ exhibits lower motion intensity but offers higher video quality, more diverse backgrounds, and greater ethnic diversity.

To verify the importance of motion intensity in training videos, we conducted comparative experiments. 
We selected an equal number of high and low-motion intensity videos from the LiveBroadcast dataset to match the size of the VFHQ dataset, creating LiveBroadcast (high) and LiveBroadcast (low) subsets. 
The results displayed in Table~\ref{tab:motion} indicate that training on LiveBroadcast (high) outperforms training on VFHQ and LiveBroadcast (low). 
Joint training on both the VFHQ and LiveBroadcast datasets yields the highest reconstruction performance.

\vspace{-0.2cm}

\begin{table}[t]
\centering
\tabcolsep=3pt
\begin{tabular}{cccc}
\toprule
             \textbf{Training Datasets} & PSNR$\uparrow$  & SSIM$\uparrow$ & FVD$\downarrow$ \\ \hline
VFHQ &29.54 &0.9010 &62.51  \\ 
LiveBroadcast (low) &29.29 &0.8986 &69.36  \\ 
LiveBroadcast (high) &29.60 &0.9008 &55.84  \\ 
\rowcolor[HTML]{E6E6E6}VFHQ+LiveBroadcast  &29.92 &0.9054 &49.11   \\ \bottomrule
\end{tabular}
\caption{Quantitative analysis of different training datasets for video reconstruction in Stage \Rmnum{1}.}
\label{tab:motion}
\end{table}

\begin{table}[t]
\centering
\tabcolsep=3pt
\begin{tabular}{ccccc}
\toprule
             \textbf{Degradation} & PSNR$\uparrow$  & SSIM$\uparrow$ & FVD$\downarrow$ & IDS$\uparrow$ \\ \hline
Rand &26.92 &0.8424 &156.2 &0.9120 \\ 
\rowcolor[HTML]{E6E6E6}Consistent   &27.47 &0.8641 &105.1 &0.9312   \\ \bottomrule
\end{tabular}
\caption{Quantitative analysis for different degradation strategies.}
\label{tab:quan1}
\end{table}

\section{Quantitative Analysis}

\subsection{Degradation: Consistent v.s. Stochastic}
In Stage \Rmnum{2}, our method applies a consistent degradation approach, where all frames in a training video exhibit consistent degradation parameters: Gaussian kernel $k_\sigma$, Gaussian noise $n_\delta$ and constant rate factor $r$. 
We explored an alternative approach using a stochastic degradation strategy, in which the degree of degradation for each frame is randomized. 
This stochastic setting disrupts the coherence between frames, significantly increasing the model's learning difficulty and introducing instability during training. 
As demonstrated in Table~\ref{tab:quan1}, the consistent degradation strategy outperforms the stochastic setting.

\subsection{Training: Incremental v.s. Holistic}
\begin{figure}[t]
    \centering
    \includegraphics[width=1.0\linewidth]{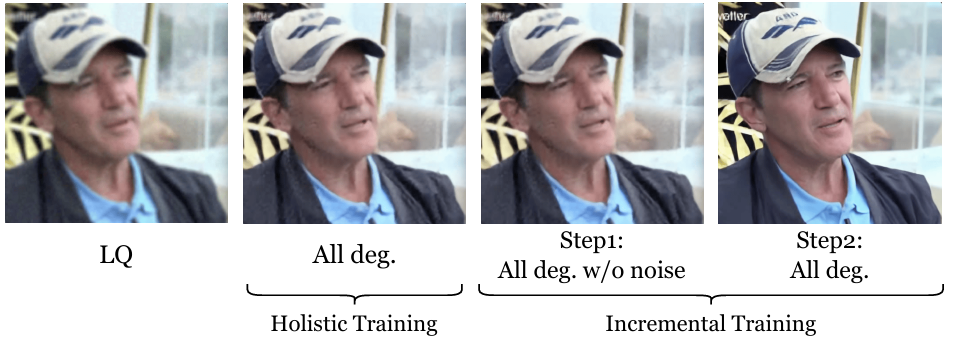}
    \caption{Visualization of restoration results when applying different training strategies in Stage \Rmnum{2}.}
    \label{fig:aba_vis2}
\end{figure}

In Stage \Rmnum{2}, directly using fully degraded videos for model training, referred to as holistic training, results in a substantial number of noise artifacts in the restoration outputs, as illustrated by the second image in Figure~\ref{fig:aba_vis2}. 
We hypothesize that stronger Gaussian blur diminishes the visibility of Gaussian noise, thereby reducing the model's de-noising effectiveness. 
To address this, we propose an innovative incremental training strategy. 
Initially, the model's de-blurring capability is developed using noise-free degraded training videos (Step 1). 
Subsequently, fully degraded videos are employed to enhance the model's blind face restoration ability (Step 2). 
As demonstrated by the restoration output in the fourth image of Figure~\ref{fig:aba_vis2} and the experimental results in Table~\ref{tab:quan2}, the incremental training strategy effectively harnesses the model's potential, leading to improved restoration capabilities. 

\begin{table}[t]
\centering
\tabcolsep=3pt
\begin{tabular}{ccccc}
\toprule
             \textbf{Strategy} & PSNR$\uparrow$  & SSIM$\uparrow$ & FVD$\downarrow$ & IDS$\uparrow$ \\ \hline
Holistic &27.05 &0.8486 &143.5 &0.9233 \\ 
\rowcolor[HTML]{E6E6E6}Incremental   &27.47 &0.8641 &105.1 &0.9312   \\ \bottomrule
\end{tabular}
\caption{Quantitative analysis for different training strategies.}
\label{tab:quan2}
\end{table}

\subsection{Training Resolution: 512$^2$ v.s. 256$^2$}

\begin{figure}[t]
    \centering
    \includegraphics[width=1.0\linewidth]{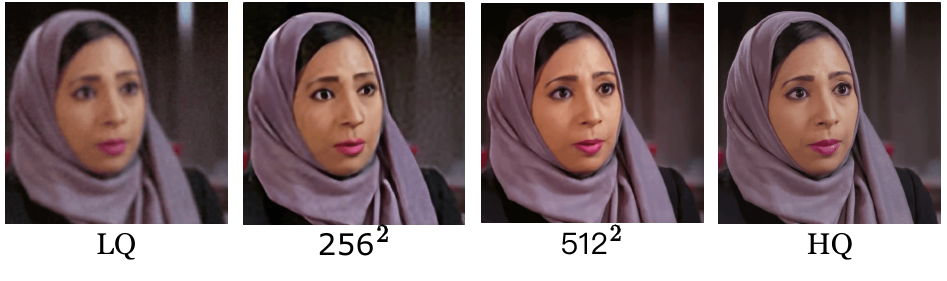}
    \caption{Visualization of restoration results when training with different resolution videos.}
    \label{fig:aba_vis1}
\end{figure}

In Stage II of training the lookup transformers and the low-quality encoder using LQ-HQ video pairs, we experimented with videos at various resolutions, specifically 512$^2$ and 256$^2$. 
Due to the presence of learnable position embeddings, the output video resolution must match that of the training videos in Stage \Rmnum{2}. 
Although using a lower resolution enhances training efficiency, it adversely affects the quality and granularity of the input, and limits the expressiveness of the output. 
As demonstrated in Figure~\ref{fig:aba_vis1} and Table~\ref{tab:quan3}, training with videos at a resolution of 512$^2$ proves to be more effective in Stage \Rmnum{2} compared to 256$^2$. 

\begin{table}[t]
\centering
\tabcolsep=3pt
\begin{tabular}{ccccc}
\toprule
             \textbf{Resolution} & PSNR$\uparrow$  & SSIM$\uparrow$ & FVD$\downarrow$ & IDS$\uparrow$ \\ \hline
256$^2$ &26.43 &0.8139 &181.5 &0.8436 \\ 
\rowcolor[HTML]{E6E6E6}512$^2$   &27.47 &0.8641 &105.1 &0.9312  \\ \bottomrule
\end{tabular}
\caption{Quantitative analysis for different resolutions of training videos in Stage \Rmnum{2}.}
\label{tab:quan3}
\end{table}

\vspace{-0.2cm}

\begin{figure*}[t]
    \centering
    \includegraphics[width=0.9\textwidth]{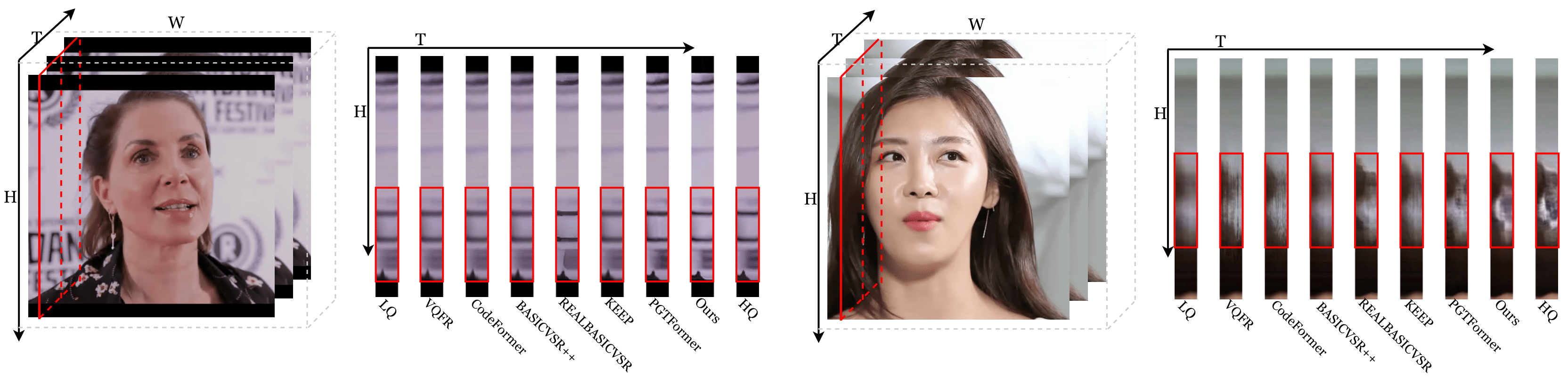}
    \caption{
    Comparison of temporal profile on the VFHQ-Test for BFVR task. We select a column to observe the changes across time. 
    }
    \label{fig:bfvr_vis3}
\end{figure*}

\begin{figure*}[t]
    \centering
    \includegraphics[width=0.9\textwidth]{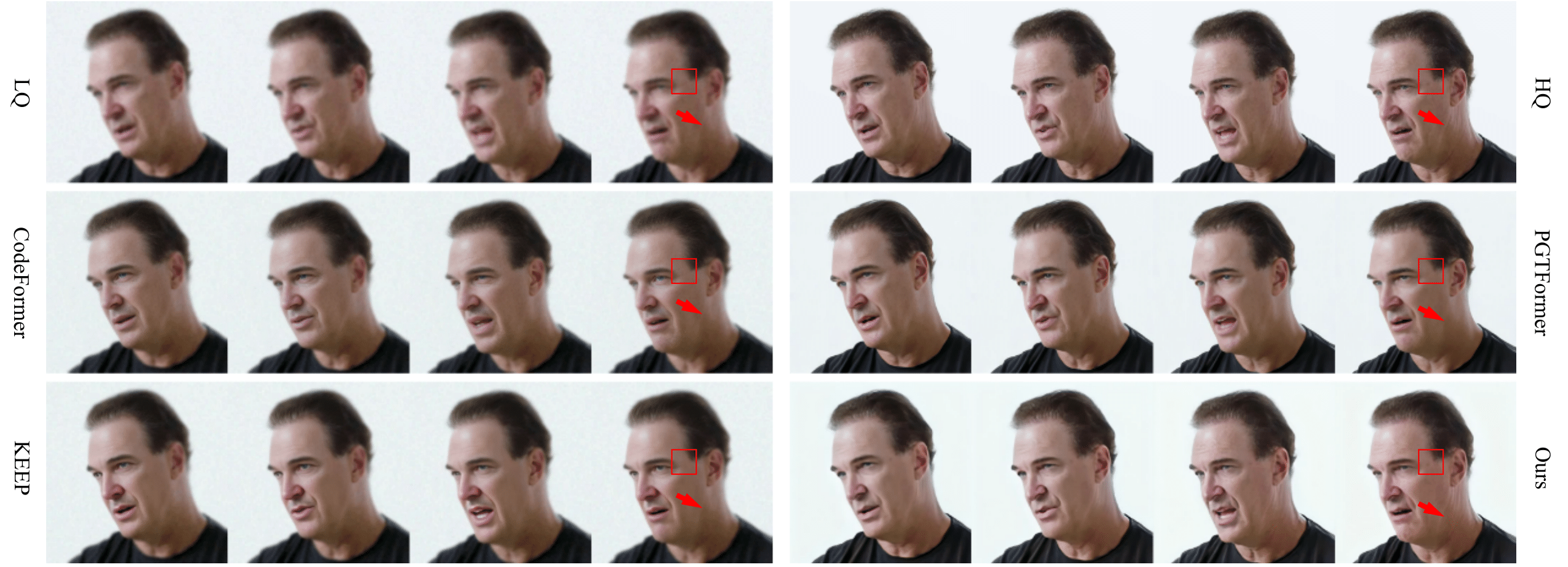}
    \caption{Qualitative comparison on the VFHQ-Test for BFVR task. 
    Our method achieves higher fidelity and demonstrates fewer hallucination cases than other methods, such as freckles in the red box and the wrinkles indicated by the arrow.
    }
    \label{fig:bfvr_vis2}
\end{figure*}

\section{Qualitative Analysis}

\subsection{Analysis of codebook usage}

\begin{figure}[t]
    \centering
    \includegraphics[width=1.0\linewidth]{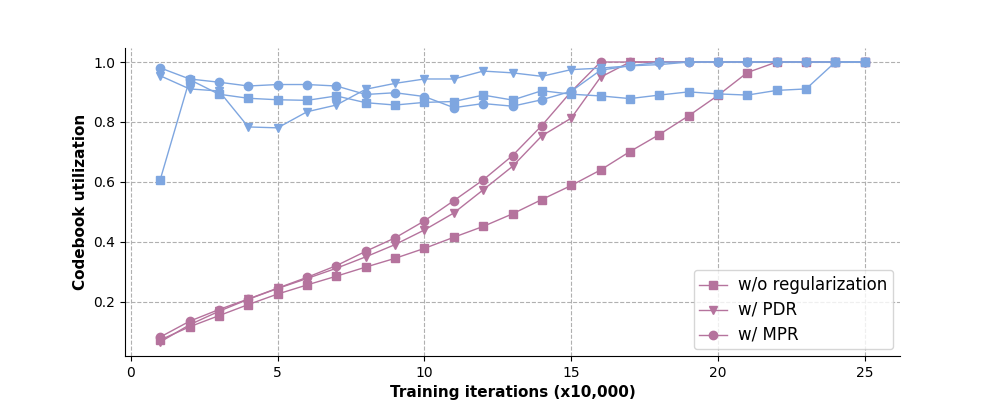}
    \caption{Visualization of the changes in codebook utilization across training iterations.}
    \label{fig:cb_training}
\end{figure}

Figure~\ref{fig:cb_training} illustrates the variation in codebook utilization across training iterations, demonstrating that MPR enhances the convergence speed of codebooks. 
The convergence curves for the spatial and temporal codebooks differ significantly due to the substantial differences in the values of code items within these codebooks. 
Compared to the PDR variant, the proposed MPR strategy facilitates a more fair and efficient updating of code items, thereby accelerating codebook learning.

\vspace{-0.2cm}


\subsection{More visualization for BFVR task}
To facilitate a more effective comparison of restoration results across different methods for the BFVR task, we select a column from each frame and stack them along the temporal dimension to visualize changes over time.
As shown in Figure~\ref{fig:bfvr_vis3}, our approach exhibits superior temporal consistency, evident by temporal textures more closely resembling those of the high-quality original videos. 
Additionally, in Figure~\ref{fig:bfvr_vis2}, we present four frames, with a gap of six frames between each, from one test video for the BFVR task. 
Methods such as CodeFormer, KEEP, and PGTFormer fail to preserve facial components faithfully, particularly freckles highlighted in the red box and wrinkles indicated by the arrow. 
In contrast, our method achieves better spatial consistency by effectively preserving portrait features in low-quality inputs while also maintaining inter-frame portrait feature consistency.

\section{Limitations and Future Work}

\begin{figure}[t]
    \centering
    \includegraphics[width=1.0\linewidth]{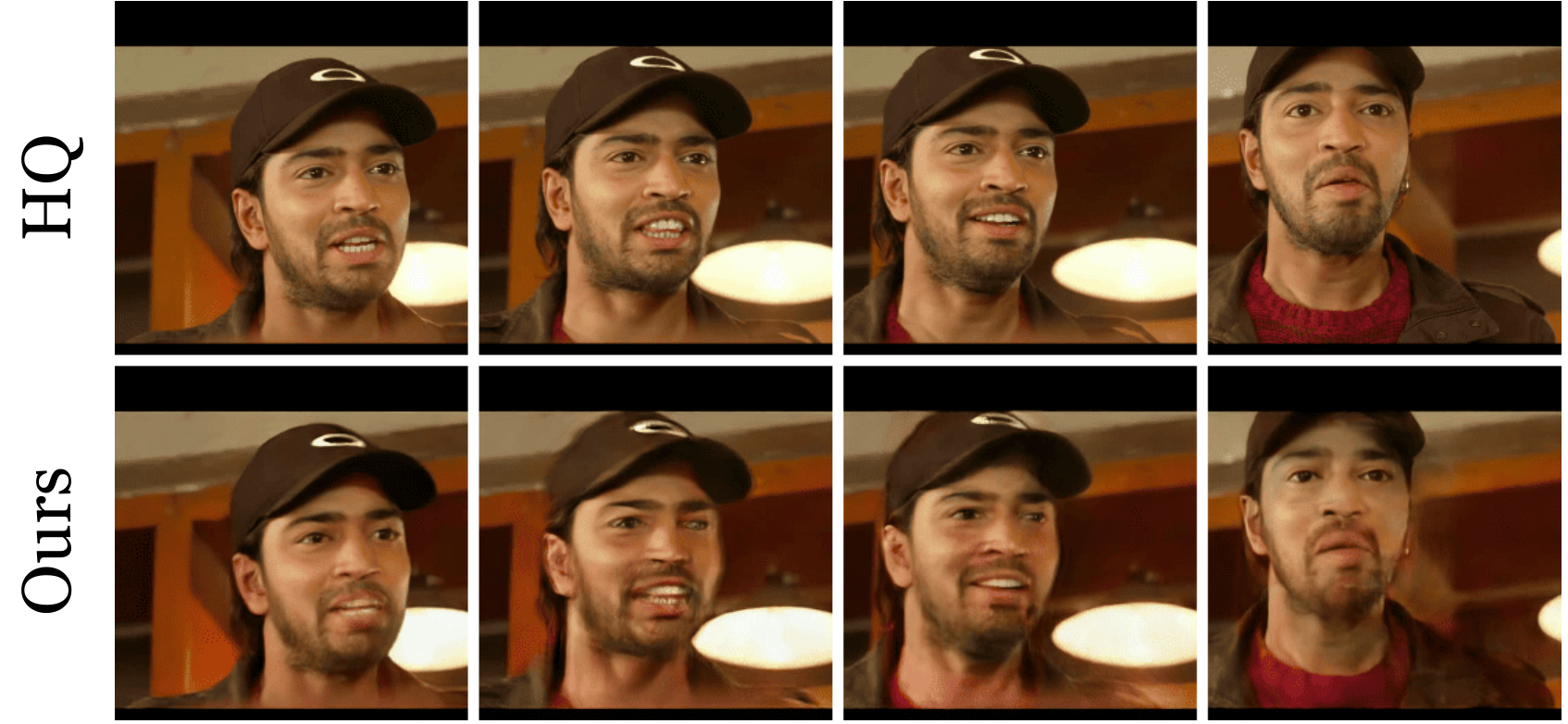}
    \caption{Visualization of a failure case generated by our method.}
    \label{fig:bad_case}
\end{figure}

Figure~\ref{fig:bad_case} illustrates a failure case in which our method may introduce video blurring. 
This occurs when the character's motion is excessively dynamic or when camera switching leads to discontinuities in facial features between frames. 
To address these issues, we intend to incorporate more sophisticated temporal modules, such as a local temporal attention mechanism. 
Additionally, we aim to enhance the spatio-temporal compression rate of 3DVAE to reduce training consumption and improve inference efficiency. 
Furthermore, we plan to integrate temporally degraded data into the BFVR task training, combining the restoration model with a de-flickering module to develop a more generalized video face enhancer.

\end{document}